\documentclass[manuscript,screen]{acmart}

\usepackage{algorithm}
\usepackage{algorithmic}
\usepackage{booktabs}
\usepackage{multirow}
\usepackage{amsmath}
\usepackage{amsthm}
\usepackage{lineno}
\usepackage{breakurl}
\usepackage{tabu}
\usepackage{tabularx}
\usepackage[caption=false,font=footnotesize]{subfig}
\usepackage{makecell}  

\usepackage{colortbl}
\usepackage{bm}  
\definecolor{my_gray}{RGB}{230,230,230} 

\AtBeginDocument{%
  }

\begin{document}

\title{FedCD: A Fairness-aware Federated Cognitive Diagnosis Framework}


\author{Shangshang Yang}
\email{yangshang0308@gmail.com}
\orcid{0000-0003-0837-5424}
\affiliation{%
  \institution{Anhui University}
  \city{Hefei}
  \state{Anhui}
  \country{China}
}

\author{Jialin Han}
\email{wa23201025@stu.ahu.edu.cn}
\orcid{0009-0007-1774-0979}
\affiliation{%
  \institution{Anhui University}
  \city{Hefei}
  \state{Anhui}
  \country{China}
  }

\author{Xiaoshan Yu}
\email{yxsleo@gmail.com}
\orcid{0000-0003-3728-6914}
\affiliation{%
  \institution{Anhui University}
  \city{Hefei}
  \state{Anhui}
  \country{China}
}

\author{Ziwen Wang}
\email{wzw12sir@gmail.com}
\orcid{0009-0005-1552-3976}
\affiliation{%
  \institution{Anhui University}
  \city{Hefei}
  \state{Anhui}
  \country{China}
}

\author{Hao Jiang}
\email{haojiang@ahu.edu.cn}
\orcid{0009-0001-8534-4130}
\affiliation{%
  \institution{Anhui University}
  \city{Hefei}
  \state{Anhui}
  \country{China}
  }

\author{Haiping Ma}
\email{hpma@ahu.edu.cn}
\orcid{0000-0002-3115-6855}
\affiliation{%
  \institution{Anhui University}
  \city{Hefei}
  \state{Anhui}
  \country{China}
}
\authornote{Corresponding Author.}

\author{Xingyi Zhang}
\email{xyzhanghust@gmail.com}
\orcid{0000-0002-5052-000X}
\affiliation{%
  \institution{Anhui University}
  \city{Hefei}
  \state{Anhui}
  \country{China}
}

\author{Geyong Min}
\email{G.Min@exeter.ac.uk}
\orcid{0000-0003-1395-7314}
\affiliation{%
  \institution{University of Exeter}
  \city{Exeter}
  \state{Devon}
  \country{UK}
  }



\renewcommand{\shortauthors}{Shangshang Yang et al.}

\begin{abstract}
  With the rapid development of Internet technologies, online intelligent education platforms have generated a vast amount of distributed student learning data. 
This influx of data presents opportunities for cognitive diagnosis (CD) to assess students' mastery of knowledge concepts while also raising significant data privacy and security challenges.
To cope with this issue, federated learning (FL) becomes a promising solution by jointly training models across multiple local clients without sharing their original data.
However, the data quality problem, caused by the ability differences and educational context differences between different groups/schools of students, further poses a challenge to the client fairness of models.
To address this challenge, this paper proposes a fairness-aware \textbf{fed}erated \textbf{c}ognitive \textbf{d}iagnosis framework (FedCD) 
to jointly train CD models  built upon a novel parameter decoupling-based personalization strategy, preserving privacy of data and achieving precise and fair diagnosis of students on each client.
As an FL paradigm,  FedCD trains a local CD model for the students in each client
 based on its  local student learning data, 
 and each client uploads its partial model parameters to the central server for parameter aggregation according to the devised innovative personalization strategy.
 The main idea of this  strategy is to decouple model parameters into two parts: the first   is used as  locally personalized parameters, containing diagnostic function-related model parameters,  to diagnose each client’s students fairly;
 the second is the globally shared parameters across clients and the server, containing exercise embedding parameters, which are updated via fairness-aware aggregation, to alleviate inter-school unfairness.
  Experiments on three real-world datasets demonstrate the
effectiveness of the proposed FedCD framework and the personalization strategy compared to five FL approaches under three CD models. 
\end{abstract}

\ccsdesc[500]{Information systems~Intelligent education systems}
\ccsdesc[500]{ Applied computing~E-learning}
\ccsdesc[500]{Computing methodologies~Federated learning}

\keywords{Intelligent education, educational data mining, learner modeling,  cognitive diagnosis, fairness-aware learning, federated learning}

\received{1 Junly 2025}

\maketitle

\section{Introduction}
\begin{figure*}[t]
		\centering
  		\subfloat[The CD  process]{\includegraphics[width=0.17\linewidth]{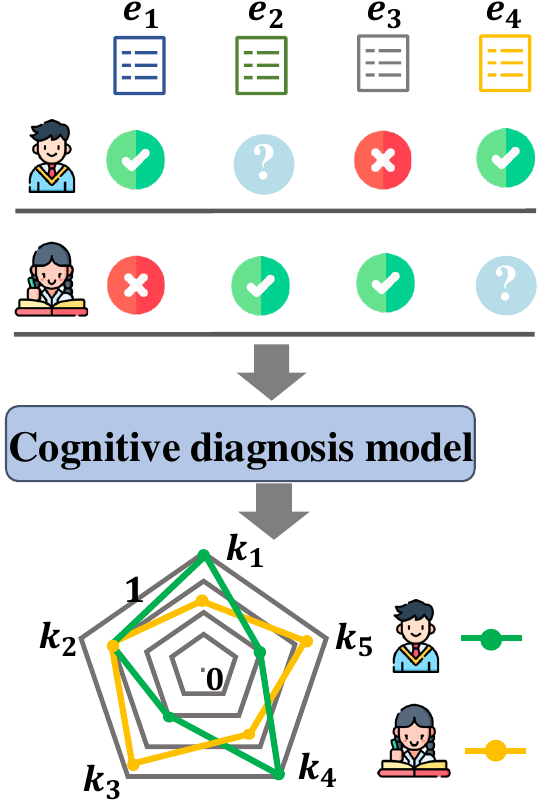}}
      	\subfloat[The client fairness issue for FL in  education]{\includegraphics[width=0.40\linewidth]{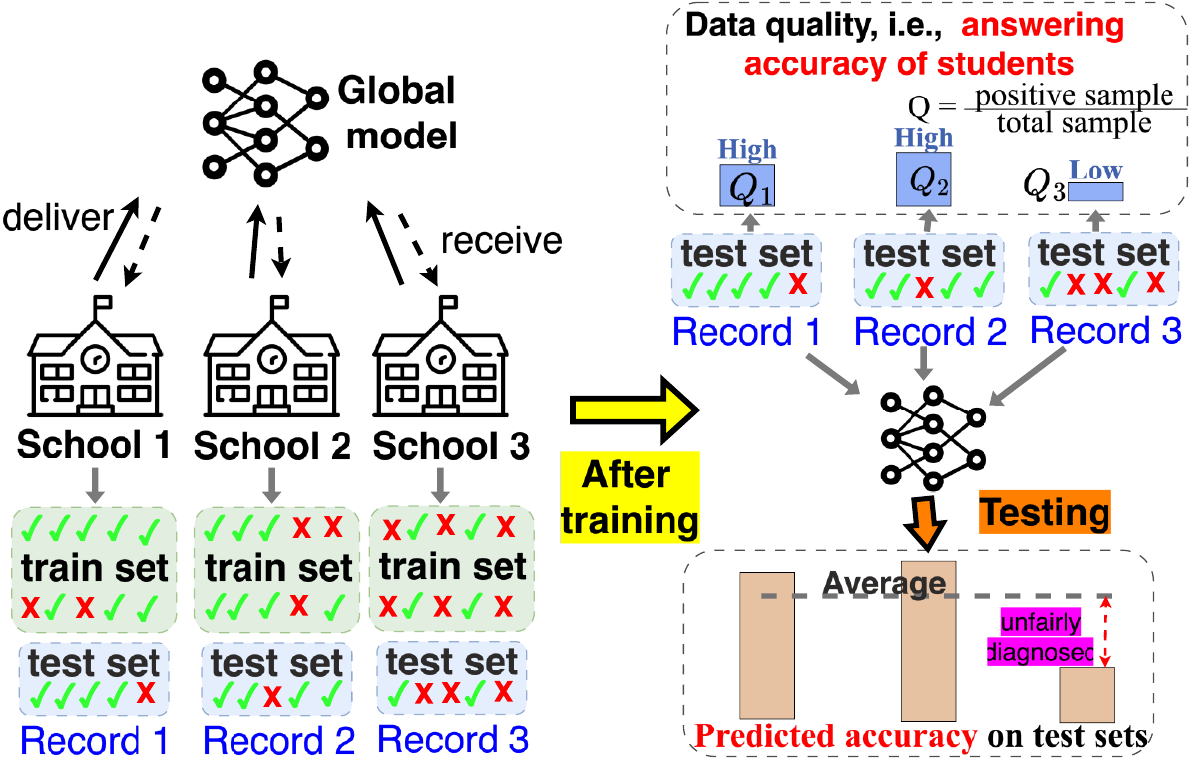}}
          \subfloat[Case study of the  fairness issue]{\includegraphics[width=0.42\linewidth]{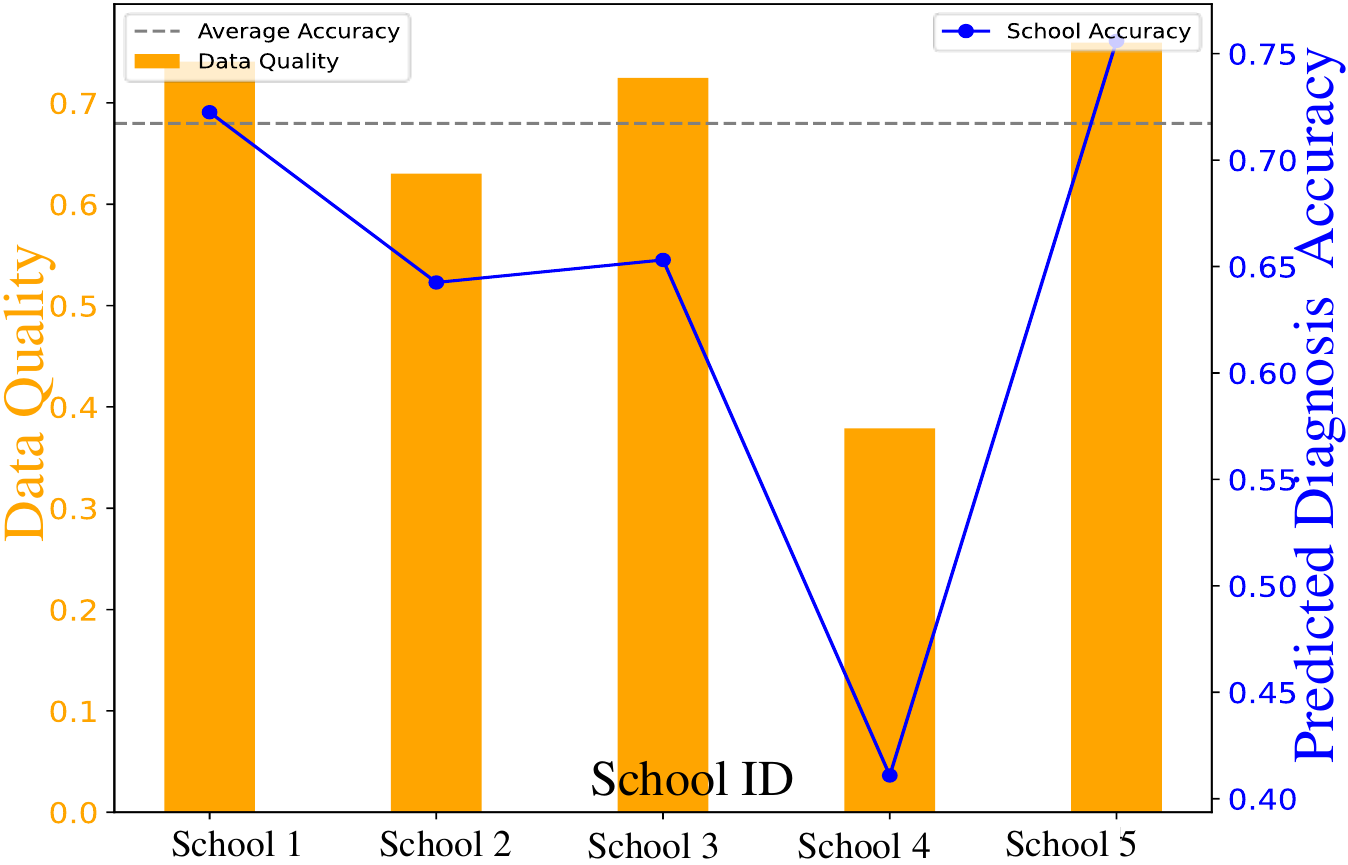}}

		\caption{(a) A toy example of CD. 
        (b)  The client fairness issue: common FL approaches fail to fairly diagnose different schools of students due to the poor quality of student learning data, i.e., low answering accuracy of students, in some schools, such as School 3. (c)  A case study by implementing FedAvg and NCD on five schools of students within the ASSISTment2009 dataset. }
    \label{fig1}

\end{figure*}

In recent years, many online intelligent education platforms (OIDPs) have emerged with significant growth,  such as JunYi~\cite{hochreiter1998vanishing}, Khan~\cite{Khan}, Coursera~\cite{Coursera}, MOOC~\cite{MOOC}, and PTA \cite{hu2023ptadisc}. 
These platforms have attracted many users (i.e., students or learners) and have generated massive amounts of various learning data~\cite{pu2024elakt}.
By collecting and analyzing these distributed learning data, 
OIDPs can assess the abilities of users~\cite{liu2024fdkt} and then provide them effective learning recommendation services~\cite{Tung2025Incorporating}, including exercise recommendation~\cite{yang2023cognitive}, course recommendation~\cite{courserecommendation}, and learning path recommendation~\cite{yang2023evolutionaryGCD}, to further enhance their abilities on poorly mastered knowledge concepts.
The process of diagnosing students' mastery on specific knowledge concepts is called cognitive diagnosis (CD)~\cite{wang2020neural}, which is a fundamental upstream technique but plays a pivotal role in OIDPs because the accuracy of the diagnosis results profoundly influences the downstream recommendation quality~\cite{gao2021rcd}.
For better understanding, Fig.~\ref{fig1}(a) gives a toy example of CD:
two students' knowledge mastery of five knowledge concepts is obtained by modeling their answer/response prediction.

To model these collected historical learning data of users, 
researchers have developed a series of cognitive diagnosis models (CDMs) based on various ideas~\cite{wang2024survey}, which can be categorized into three types.
The first type focuses on the model interpretability and thus 
creates CDMs based on the theory of educational psychology~\cite{wang2020neural}, where 
IRT \cite{embretson2013item},  MIRT~\cite{reckase2009multidimensional}, and  DINA~\cite{Torre2009DINA} are the representative approaches. 
Considering the powerful modeling ability of neural networks (NNs)~\cite{alom2018history},
the second genre aims to incorporate NNs with the above CDMs to enhance diagnosis accuracy, 
where the representatives include NCD~\cite{wang2020neural}, KaNCD~\cite{wang2022kancd}, KSCD~\cite{ma2022knowledge} and RCD \cite{gao2021rcd}.
To save human labor on manually designing CDMs, the third leverages the idea of automated machine learning~\cite{hutter2019automated} like neural architecture search~\cite{zoph2016neural,yang2022accelerating,sun2019evolving} to search CDMs, 
with  EMO-NAS-CD~\cite{yang2024evolutionaryCD,yang2023designing} and NAS-GCD~\cite{yang2023evolutionaryGCD} as the representative.

Although  CDMs achieved considerable success, they rely on collecting all users’ learning data together for centralized training~\cite{zhu2021federated}. 
The existing CDMs are effective in terms of model training but have been criticized for data privacy and security risks~\cite{zhang2021survey}, because user data is often highly sensitive but centralized storage and processing raise privacy concerns and increase data leakage risks~\cite{Liu2022Fairness}. 
Therefore, many users are reluctant to share their data, limiting the practicality of the current paradigm of CDMs~\cite{wu2021federated}.
To achieve the data privacy-preserved cognitive diagnosis paradigm,  
an intuitive idea is to leverage the federated learning (FL)~\cite{jin2023federated} for CD. 
FL enables users to keep their data on local clients without sharing it 
and trains models by only communicating model parameters between user clients and the central server. 
FL has made significant progress in many tasks~\cite{li2020review}, 
such as recommendation~\cite{yang2020federated}, medical image~\cite{adnan2022federated}
and Internet of Things~\cite{nguyen2021federated}, but has not been explored in CD to protect user privacy.

To fill this gap, this paper seeks to integrate FL  with CD to ensure the diagnosis accuracy of CDMs while preserving user privacy. 
{However, the non-independent and identically distributed (Non-IID) problem~\cite{zhu2021federated} in FL will likely become more pronounced when applied to CD, which is mainly raised by the data quality problem among different schools of students. Specifically, there is a significant gap in the abilities of students in different levels of schools, and thus the quality of the collected students' learning data is substantially different, i.e., there are essential data quality differences among different schools of students.}
As shown in Fig.~\ref{fig1}(b),  students at different schools often exhibit significant ability and education context disparities, such as school course schedules, available resources, and their initial ability levels at admission, where students at School 3 are much poorer than the other two. 
 As a result, CDMs obtained by common FL approaches may perform unfairly on some clients of students who are out of the distribution, i.e., fail to diagnose accurately at School 3 in Fig.~\ref{fig1}(b).
 To validate the client fairness issue, 
 an easy case study implementing  NCD with FedAvg~\cite{li2019convergence} is performed on five schools of students in ASSISTment2009~\cite{Assistments09}. 
 In Fig.~\ref{fig1}(c),    students at School 4 hold lower answering accuracy (i.e., poorer abilities)  than others,  and the NCD model trained by FedAvg showcases much poorer diagnosis accuracy on them than others. 
 This indicates that common FL approaches indeed encounter the client fairness issue when applied to CD.
{Although HPFL~\cite{hpfl} and AHPFL~\cite{liu2023federated} employ the FL technique for CD, 
it aims to solve the heterogeneous data within clients but not to address the 
client fairness issue, leading to limited scalability.}

 To address the above issues, we propose a fairness-aware federated cognitive diagnosis framework (FedCD) 
 to preserve the privacy of student learning data and accurately diagnose students in each client. 
 Specifically,  our  contributions are  summarized below:
  \begin{itemize}

     \item 
     To tackle the client fairness issue,  a novel parameter decoupling-based personalization strategy is devised for the proposed  FedCD to fairly and accurately diagnose students at any client regardless of the client's data quality.
      Following the fundamental procedures of common FL approaches, the proposed FedCD trains a local CDM for each client using local student learning data and  aggregates  model parameters in the central server to update parameters and assign them to clients, repeating multiple rounds.

     \item  
     The proposed parameter decoupling-based personalization strategy divides the model parameters in each client into two parts: a locally personalized part and a globally shared part across all clients and the server. The former contains the diagnostic function-related model parameters to capture the local student characteristics.  The latter contains exercise embedding parameters, 
     updated by fairness-aware aggregation, which was designed to mitigate the school fairness issue.

     \item    Extensive experimental results demonstrate the effectiveness of the proposed FedCD compared to five FL approaches under three CDMs  and the effectiveness of the new personalization strategy. 
     Besides, the FedCD is validated to be effective in solving the client fairness issue to achieve fair diagnosis.

\vspace{-3mm}
 \end{itemize}

\section{Related Work}

\subsection{Cognitive Diagnosis}

As one of many user modeling techniques~\cite{huang2020learning}, cognitive diagnosis aims to extract students’ features from their large amounts of historical learning data, similar to recommender systems~\cite{gurbanov2016modeling}. However, due to the nature of the CD task, the extracted features must be interpretable. Thus, existing CDMs represent each student’s features through a single ability scalar or a latent vector that indicates their mastery level of each knowledge concept. As two pioneering approaches, IRT~\cite{embretson2013item} and DINA~\cite{Torre2009DINA} represent student mastery using a continuous scalar and a binary vector, respectively, where the answer prediction process is completed using a logistic function in IRT and an inner-product operation in DINA. Inspired by IRT and DINA, later-developed CDMs adopt the idea of using a continuous vector to represent students’ knowledge proficiency or states, but these CDMs employ different techniques for diagnosis, including matrix factorization-based MF~\cite{koren2009matrix}, multilayer perceptron-based NCD~\cite{wang2020neural}, KaNCD~\cite{wang2022kancd}, KSCD~\cite{ma2022knowledge}, graph NN-based RCD~\cite{gao2021graph}, CDGK~\cite{wang2021using}, SCD~\cite{scd}, Bayesian network-based HierCDF~\cite{li2022hiercdf}, inductive learning-based ICDM~\cite{liu2024inductive}, and others~\cite{yang2024evolutionaryCD,yang2023evolutionaryGCD}. Despite the variety of techniques in existing CDMs, their effectiveness largely depends on the centralized training of models using all students’ data, raising concerns about data privacy and potential data leakage~\cite{jin2023federated}. Fortunately, a federated approach, HPFL~\cite{hpfl} (AHPFL~\cite{liu2023federated} is a extension of HPFL), uses a hierarchical local model to tackle the data heterogeneity problem. However, HPFL relies heavily on a fixed model, GUM, which limits its scalability and applicability to other CDMs. More importantly, HPFL does not address client fairness issues. Therefore, it is urgent to develop a scalable CD paradigm to ensure data privacy and fairness.

\subsection{Federated Learning}

As one of the most promising techniques, federated learning employs a central server to coordinate multiple clients in collaboratively training a model without directly accessing their local data~\cite{kairouz2021advances}. To effectively aggregate model parameters, researchers have proposed various strategies, such as FedAvg~\cite{fedavg}, FedAtt~\cite{fedatt}, FedSGD~\cite{mcmahan2017communication}, and LoAdaBoost~\cite{huang2020loadaboost}. Based on these fundamental FL approaches, federated learning has been widely extended to various domains and has achieved great success, including banking~\cite{yang2019ffd}, healthcare~\cite{adnan2022federated}, personal devices~\cite{yang2018applied}, and many others~\cite{li2020review}. Among these, FL-based recommendation~\cite{zhang2023dual,zhang2024gpfedrec}, also known as federated recommendation (FedRec), is most closely related to cognitive diagnosis. The representative FedRec approaches include MetaMF~\cite{lin2020meta}, which uses a meta-network to generate private item embeddings and a rating prediction model; FedNCF~\cite{perifanis2022federated}, which applies the classical NCF~\cite{he2017neural} approach to the federated setting; FedPerGNN~\cite{wu2022federated}, which learns high-order user-item information through graph NNs; and PFedRec~\cite{zhang2023dual}, which incorporates a parameter personalization mechanism. Inspired by their success, this paper aims to apply FL to CD to train effective CDMs while preserving students’ learning data privacy. However, the Non-IID problem is more pronounced in the intelligent education context, as the differences in data quality are more significant than in recommendation scenarios, according to~\cite{wu2021federated}, where different schools or groups of students generally exhibit considerable ability differences. This type of data quality issue can lead to unfair diagnoses among different student clients when using standard FL approaches. Although some fair FL approaches have been proposed~\cite{Liu2022Fairness,yu2020fairness}, they are mainly tailored for specific tasks and may not be suitable for CD. Therefore, a fairness-aware federated cognitive diagnosis framework is needed.

\subsection{Difference between Federated Recommendation and Federated Cognitive Diagnosis}

Federated recommendation is an emerging paradigm that enables collaborative filtering and personalized content delivery while preserving user privacy through decentralized learning. 
It typically models both users and items to infer users' preferences for unseen items, aiming to predict the likelihood of user-item interactions  ~\cite{chai2020secure}.
While both CD and recommendation systems utilize latent user representations, they differ substantially in data structure, objectives, and modeling paradigms. CD involves a multi-entity structure, i.e. students, exercises, knowledge concepts, and schools (as clients)~\cite{hpfl}, leading to intra-client data imbalance, a challenge not present in typical recommendation systems where each client represents a single user. CD aims to assess student mastery over specific concepts using interpretable models aligned with a Q-matrix and domain constraints, such as monotonicity~\cite{zhang2024towards}. 
In contrast, recommendation focuses on predicting user-item preferences without interpretability or concept alignment~\cite{lin2020fedrec}. Furthermore, CD functions as a core diagnostic task in educational systems, supporting downstream applications like personalized learning paths~\cite{li2023graph}. Existing federated recommendation methods often underperform in CD settings due to these fundamental differences~\cite{hpfl}, making it inappropriate to treat CD as a subset of recommendation.

\section{Preliminaries}

\subsection{Cognitive Diagnosis}
For the CD task in one online intelligent education  platform, 
 three sets of items are generally collected together for prediction modeling, 
 including a set of $M$ exercises $\mathcal{E} = \{ e_1, e_2, \ldots, e_M \}$,
a set of $N$ students $\mathcal{S} = \{ s_1, s_2, \ldots, s_N \}$, 
and a set of $K$ knowledge concepts (concepts for short) $ \mathcal{C} = \{ c_1, c_2, \ldots, c_K \}$. 
In addition, a prior exercise-concept relation  matrix (called Q-matrix) is also needed, denoted as $Q=(Q_{jk}\in\{0,1\})^{M\times K}$, where $Q_{jk}=1$ refers to  the concept $c_k$  not included in the exercise $e_j$  and  $Q_{jk}=0$ otherwise.

In traditional centralized training of CDMs, all students' learning data (known as exercising response logs) 
need to be collected together as a set of triplets $\mathcal{R}=\{(s_i,e_j,r_{ij})|s_i\in \mathcal{S}, e_j\in \mathcal{E}, r_{ij}\in\{0,1\}\}$. Here $r_{ij}$ refers to the response/answer of student $s_i$ on exercise $e_j$. $r_{ij}=1$ means the answer is correct and $r_{ij}=0$ otherwise. 
As a result,  \textbf{given} students' response logs $\mathcal{R}$ and the Q-matrix, 
the \textbf{goal} of CD is to  reveal students'  proficiency on concepts by training  a CDM $\mathcal{M}(\cdot|\theta)$ to model the students'  answer prediction as follows:
\begin{equation}	\label{eq:embed}
\left.
\begin{aligned}
&\mathop{\min}\limits_{\theta\in \Theta} \sum_{(s_i,e_j,r_{ij})\in \mathcal{R}}Loss\left(r_{ij}, \mathcal{M}(\mathbf{x}_i^S, \mathbf{x}_j^E|\theta)\right)\\
\mathcal{M}(\mathbf{x}_i^S, \mathbf{x}_j^E|\theta) =& 
	\left\{
	\begin{aligned}
		\mathbf{h}_S = Emb_1(&\mathbf{x}_i^S|\theta^S) ,\  \mathbf{h}_E = Emb_2(\mathbf{x}_j^E|\theta^E)\\
  \mathbf{h}_C = \mathbf{x}_j^E \times Q &= (Q_{j1}, Q_{j2},\cdots, Q_{jK})\\
  \hat{r}_{ij} = &\mathcal{I}(\mathbf{h}_S,\mathbf{h}_E,\mathbf{h}_C|\theta^I)\\
	\end{aligned}
\right.\\
\end{aligned}\right..
\end{equation}
Here $\mathbf{x}_i^S \in \{0,1\}^{1\times N}$ and   $\mathbf{x}_j^E \in \{0,1\}^{1\times M}$
are one-hot vectors for student $s_i$ and exercise $e_j$, respectively. The student-related  vector $\mathbf{h}_S\in \mathbb{R}^{1\times D}$ and 
exercise-related  vector $\mathbf{h}_E\in \mathbb{R}^{1\times D}$ are  obtained by two embedding modules $Emb_1(\cdot|\theta^S)$ and $Emb_2(\cdot|\theta^E)$, parameterized by $\theta^S$ and $\theta^E$.
$\mathcal{I}(\cdot|\theta^I)$ denotes the diagnostic function to obtain the predicted answer $\hat{r}_{ij}$ of student $s_i$ and exercise $e_j$, which is parameterized by $\theta^I$.
As a result,  the model parameters $\theta$ of $\mathcal{M}(\cdot|\theta)$ contains three components, i.e., 
$\theta=\{\theta^S, \theta^E, \theta^I\}$.

After optimizing $\theta$ to minimize the loss in the first item in Eq.(\ref{eq:embed}),  
 $\mathbf{h}_S$ can be used to represent the student's knowledge proficiency, which is the diagnosis result of student $s_i$.

\subsection{Federated Learning}
Different from previous CDMs, FL learns a global model parameterized by $\bar{\theta}$ to serve all clients of users without directly accessing their data.   
As a result, the target of a common FL approach is to minimize the  loss of all $T$ clients as follows:
\begin{equation}	
\label{eq:FL}
    \min_{\bar{\theta} \in \Theta} \sum_{t=1}^{T} \alpha_{t} \mathcal{L}\left(D_t|\bar{\theta}\right),
\end{equation}
where $\mathcal{L}\left(D_t|\bar{\theta}\right)$ denotes the supervised loss on the $t$-th client  with its dataset $Dt$.
$\alpha_{t}$ is the weight for  the $t$-th client's loss. 
Taking FedAvg~\cite{fedavg} as an example, 
its $\alpha_{t}$  is defined as the fraction of $t$-th client’s  data size, i.e.,
$\alpha_{t}= |D_t|/\sum_{t}^{T} |D_t|$.  
After the global model is trained, it can be utilized to predict tasks for all users' clients.

\begin{figure*}[t]
    \centering
    \includegraphics[width=1.\linewidth]{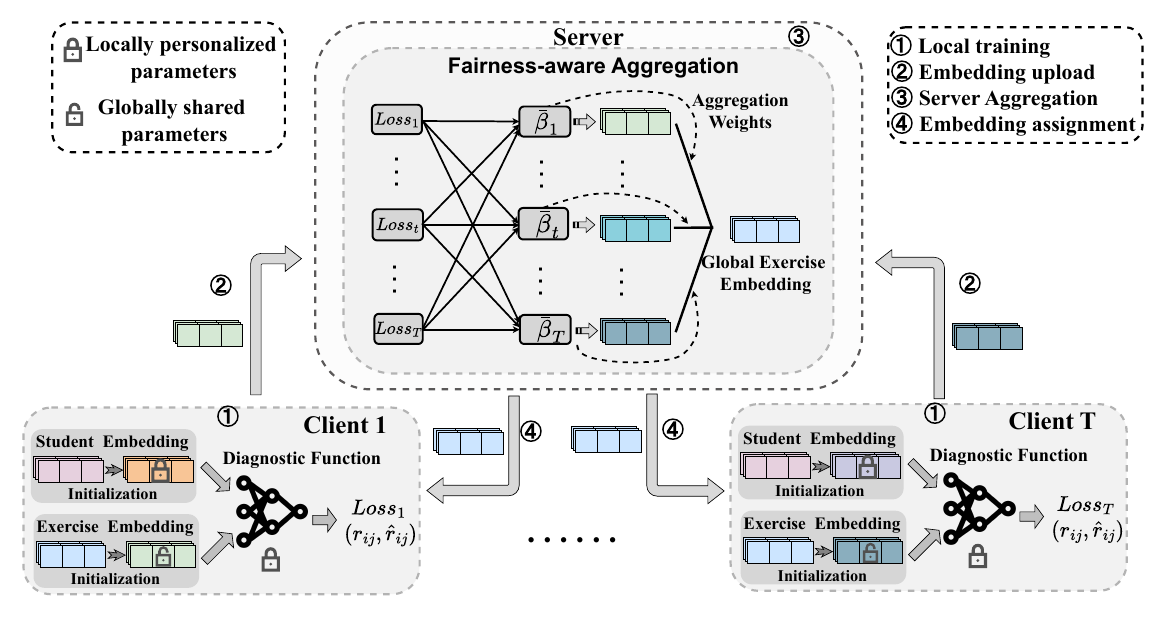} 
    \caption{
    Overview of FedCD. Each communication round consists of four steps: $\textcircled{1}$ Each school/client trains its own CDM, initializing its local exercise embedding with the globally shared embeddings. These are then substituted into the locally preserved student embedding and diagnostic function to compute the predicted loss ${Loss}_{t}(y,\hat{y})$. $\textcircled{2}$ Each client uploads its exercise embedding to the server. $\textcircled{3}$ The server determines the aggregation weights based on the chosen federated framework and aggregates the local exercise embeddings from each client to obtain the global exercise embedding. $\textcircled{4}$ Finally, the server distributes the global exercise embedding back to the clients.
}
    \label{fig2}
\end{figure*}

\subsection{Federated Cognitive Diagnosis Definition}
In  online intelligent education, suppose there are $T$ schools/groups of students $\mathcal{S}=\left\{\mathcal{S}_{T_t}=\{s_t^1, s_t^2,\cdots,s_t^{N_t}\}| 1\leq t\leq T\right\}$  
and the $t$-th school has $N_t$ students as $\mathcal{S}_{T_t}=\{s_t^1, s_t^2,\cdots,s_t^{N_t}\}$  is regarded as the $t$-th client.
Therefore, the equation $\sum_{t}^T N_t = N$ holds. 
Then, the learning data contained in $t$-th client is denoted as  $\mathcal{R}_{T_t} = \left\{(s_i,e_j,r_{ij})|s_i\in \mathcal{S}_{T_t}, e_j\in \mathcal{E}, r_{ij}\in\{0,1\}\right\}$. 
\textbf{Given} the above information, 
the goal of federated CD is to  reveal students' knowledge states by minimizing the diagnosis loss of all $T$ clients:
\begin{equation}	
        \min_{\bar{\theta} \in \Theta} \left( \sum_{t=1}^{T} \alpha_{t} \times \sum_{(s_i,e_j,r_{ij}) \in \mathcal{R}_{T_t}} \text{Loss}\left(r_{ij}, \mathcal{M}(\mathbf{x}_i^S, \mathbf{x}_j^E \mid \bar{\theta})\right) \right),
    \label{eq9}
\end{equation}
where  $\mathcal{M}(\mathbf{x}_i^S, \mathbf{x}_j^E|\bar{\theta})$ is same as that in Eq.(\ref{eq:embed}). 
Here $\bar{\theta} = \{\bar{\theta}^S, \bar{\theta}^E, \bar{\theta}^I\}$ are all the same for all clients in common FL approaches.

\section{The Proposed FedCD Framework}
\subsection{Overview}

In this section, we present the fairness-aware federated cognitive diagnosis framework. As shown in Fig.~\ref{fig2}, the client fairness issue is addressed by implementing a parameter decoupling-based personalization strategy, which separates the $t$-th client's model parameters $\theta_t=\{\theta_t^S, \theta_t^E, \theta_t^I\}$ into two parts. The first part is referred to as the \textbf{globally shared parameters}, which include the parameters of exercise embedding modules, $Emb_2$, i.e., $\{\theta_t^E\}$. The second part is known as the \textbf{locally personalized parameters}, which pertain to the parameters of the diagnostic function $\mathcal{I}(\cdot)$ and student embedding modules, i.e., $\{\theta_t^S,\theta_t^I\}$. In summary, the central server only updates the parameters $\bar{\theta}^E$ as $\bar{\theta}$ from all clients, i.e., $\bar{\theta}=\{ \bar{\theta}^E\}$, while each client retains its personalized diagnostic function parameters $\bar{\theta}^I$ and student embedding parameters $\bar{\theta}^S$, i.e., $\{{\theta_t^S, \theta_t^I}\}$. With this strategy, the procedures for each training round of the FedCD framework are as follows: First, the clients train their local models based on their own student learning data. Second, each client keeps the parameters $\theta_t^S$ and $\theta_t^I$ local for the next round of training and uploads the parameters $\theta_t^E$ to the server. Next, the server updates the exercise embedding parameters $\bar{\theta}^E$ through a simple fairness-aware weighted aggregation mechanism. Finally, the exercise embedding parameters are assigned to all clients for local model training.

In the following part, we provide a detailed introduction to the proposed FedCD. First, we explain the parameter decoupling-based personalization strategy by outlining the design of the optimization objective, client, and server, respectively. Then, we describe the learning process and summarize the overall optimization workflow in the form of an algorithm. Finally, we analyze the privacy-preserving capabilities of our approach and discuss how integrating privacy protection techniques can further enhance it.

\subsection{Parameter Decoupling-based Personalization Strategy}

 The main idea of this strategy is to maintain the parameters $\bar{\theta}^E$ as $\bar{\theta}$ in the central server by aggregating the corresponding parameters from clients, specifically the exercise embedding parameter set $\{\theta_1^E, \ldots, \theta_T^E\}$. This approach allows each client to retain locally personalized parameters $\{{\theta_t^S, \theta_t^I}\}$ for its student embedding and diagnostic function, enabling the model to address the client fairness issue and achieve fair and accurate diagnoses for each client.

 \subsubsection{Optimization Objective Design}
Due to  the locally personalized parameters of the model in each client, 
 the overall optimization objective  of FedCD will be turned to  the following:
\begin{equation}	
 \begin{aligned}
      \min_{\{\bar{\theta}^S \in \Theta^S, \bar{\theta}^E \in \Theta^E\}}  & \sum_{t=1}^{T} \alpha_{t} \times \\ 
   \sum_{(s_i,e_j,r_{ij})\in \mathcal{R}_{T_t}}&Loss\left(r_{ij}, \mathcal{M}(\mathbf{x}_i^S, \mathbf{x}_j^E| \{{\theta}_t^{S}, {\theta}_t^{E}, \theta_t^I\} )\right) ,\\
  s.t.,\ {\theta}_t^{S}, {\theta}_t^{E}, \theta_t^I = &\mathop{\arg\min}\limits_{
   \{\hat{\theta}_t^S \in \Theta^S, \hat{\theta}_t^E \in \Theta^E, \hat{\theta}_t^I\in \Theta^I\}}\\ \sum_{(s_i,e_j,r_{ij})\in \mathcal{R}_{T_t}} &Loss\left(r_{ij}, \mathcal{M}(\mathbf{x}_i^S, \mathbf{x}_j^E| \{\hat{\theta}_t^S, \hat{\theta}_t^E, \hat{\theta}_t^I\} )\right),\\
 \end{aligned}
 \label{eq:FLCD}
\end{equation}
where $\Theta^S$,  $\Theta^E$, and $\Theta^I$ denote the parameter search space.
The above equation is a bi-level optimization problem: 
 the upper-level problem  solved in the server aims to find optimal $\bar{\theta}^E$ as the initial parameters for the lower-level problem in each client, i.e., $\hat{\theta}_t^E$ takes $\bar{\theta}^E$ as initial values; 
 while the lower-level problem  aims to find optimal parameters 
$\{\theta_t^S,\theta_t^E,\theta_t^I\}$ by minimizing the loss values on the data $\mathcal{R}_{T_t}$. 
It is important to note that solving the upper-level problem requires multiple rounds. In each round, 
$\hat{\theta}_t^S$ and $\hat{\theta}_t^I$ in the lower-level problem use the $\theta_t^S$ and $\theta_t^I$  in the last  round as the initial value, while in the first round, $\hat{\theta}_t^S$ and $\hat{\theta}_t^I$ are initialized randomly.

 The loss weight of $t$-th client  $\alpha_t$ is set to 1 to consider client fairness, ignoring the data size influence.
 The forward-pass process of $\mathcal{M}(\mathbf{x}_i^S, \mathbf{x}_j^E| \{\hat{\theta}_t^S, \hat{\theta}_t^E, \hat{\theta}_t^I\} )$ is same as that in Eq.(\ref{eq:embed}), 
 which outputs the predicted answer of student $s_i$ got on exercise $e_j$, i.e., $\hat{r}_{ij}$. 
 The  binary cross-entropy loss~\cite{yang2023evolutionary} is used to measure the loss between $\hat{r}_{ij}$ and ${r}_{ij}$ as follows:
 \begin{equation}
 \label{eq_loss}
     Loss({r}_{ij},\hat{r}_{ij}) = - \left({r}_{ij} log(\hat{r}_{ij})+(1-{r}_{ij})log(1-\hat{r}_{ij}) \right).
 \end{equation}

\subsubsection{Client  Design}

The FedCD needs to train $T$ models for $T$ schools of students independently. 
These $T$ models have the same network architecture but different weight parameters, 
where the model parameters of $t$-th client are $\theta_t = \{\theta_t^S, \theta_t^E, \theta_t^I\}$.

For most  existing CDMs,  $\theta_t^S$ and $\theta_t^E$ are the same 
according to  Eq.(\ref{eq:embed}), and the student embedding module $Emb_1(\cdot)$ and the exercise embedding module $Emb_2(\cdot)$ in the $t$-th client can be denoted as
\begin{equation}
\left.
    \begin{aligned}
    \mathbf{h}_S = Emb_1(\mathbf{x}_i^S|\theta_t^S\}) = x_i^S\times\theta_t^S, \ \theta_t^S\in \mathbb{R}^{N\times D}\\
            \mathbf{h}_E = Emb_2(\mathbf{x}_j^E|\theta_t^E\}) = x_j^E\times\theta_t^E, \ \theta_t^E\in \mathbb{R}^{M\times D}\\
    \end{aligned}\right.,
\end{equation}
where $D$ is the embedding size, usually equal to $K$. 
$\theta_t^S$ and $\theta_t^E$ are two learnable parameter matrices 
to store all students' and all exercises' embedding parameters.

Generally, different CDMs have different architectures of $\mathcal{I}(\cdot)$  and thus $\theta_t^I$ is various. Here we take NCD as an example, whose forward-pass process is as follows:
\begin{equation}
    \mathcal{I}(\mathbf{h}_S,\mathbf{h}_E,\mathbf{h}_C| \theta_t^I) = \left\{
	\begin{aligned}
		&\mathbf{f}_S = \sigma(\mathbf{h}_S),\ \mathbf{f}_{diff} =                        \sigma(\mathbf{h}_E)\\ 
		&f_{disc} = \sigma(\mathbf{h}_E\times W_{disc}), W_{disc}\in                      \mathbb{R}^{D\times 1}  \\
   	    & \mathbf{y}= \mathbf{h}_C\odot(\mathbf{f}_S-\mathbf{f}_{diff} )\times          f_{disc}\\
            & \hat{r}_{ij} = \sigma( \sigma (\sigma( \mathbf{y}\times W_{fc1} )\times W_{fc2})\times W_{fc3})
	\end{aligned} 
	\right.,
\end{equation}
 where $\sigma(\cdot)$ denotes the Sigmoid activation function, 
 $\mathbf{f}_{diff} \in \mathbb{R}^{1\times D}$ and   $\mathbf{f}_{disc} \in \mathbb{R}^{1}$ represent the difficulty and the discrimination of the exercise. 
 The last item denotes the employed three-layer fully connected layers, whose three parameters are $ W_{fc1}\in \mathbb{R}^{D\times 4D}$, $ W_{fc2}\in \mathbb{R}^{4D\times 2D}$, and 
 $ W_{fc3}\in \mathbb{R}^{2D\times 1}$.
 As a result, the diagnostic function-related model parameters for $t$-th client can be represented as 
 $\theta_t^I = \{ W_{disc},W_{fc1}, W_{fc2}, W_{fc3}\}$.

As shown in Eq.(\ref{eq:FLCD}), 
equipped with the server-assigned $\bar{\theta}^E$ 
and locally-saved $\{{\theta}_t^S, {\theta}_t^I\}$ as initial $\hat{\theta}_t^E$ and $\{\hat{\theta}_t^S,\hat{\theta}_t^I\}$, each client's model needs to be trained on data $\mathcal{R}_{T_t}$
for $Num$ epochs to get its promising model parameters  $\theta_t = \{\theta_t^S, \theta_t^E, \theta_t^I\}$.
After that,  $\theta_t^E$ is uploaded to the server to generate new parameters for the next round, and $\{\theta_t^S, \theta_t^I\}$ is kept local to be used directly as the initial of the next round.

\subsubsection{Server Design}
After all clients' models are locally trained, 
the central server is responsible for aggregating their uploaded parameter sets, i.e., exercise embedding parameter set $\{\theta_1^E, \cdots, \theta_T^E\}$.

To obtain new   exercise embedding parameters $\bar{\theta}^E \in \mathbb{R}^{M\times D}$, the following fairness-aware aggregation mechanism is adopted:
\begin{equation}
\begin{aligned}
\label{eq_exer}
&\bar{\theta}^E  = \sum_{t=1}^T \left(\bar{\beta}_t \times \theta_t^E \right),\
\textrm{where}\ \bar{\beta}_t = \frac{exp(\gamma\cdot\mathcal{L}_t)}{\sum_{j=1}^T exp(\gamma\cdot\mathcal{L}_j)},  \\
&\mathcal{L}_t = \sum_{(s_i,e_j,r_{ij})\in \mathcal{R}_{T_t}}Loss\left(r_{ij}, \mathcal{M}(\mathbf{x}_i^S, \mathbf{x}_j^E| \{{\theta}_t^{S}, {\theta}_t^{E}, \theta_t^I\} )\right), \\
\end{aligned}
\end{equation}
where $\bar{\beta}_t\in \mathbb{R}^1$ is a normalized weight of $t$-th client's parameters to obtain $\bar{\theta}^E$. 
Here $\gamma$ is a parameter adjusting the importance of the client loss $\mathcal{L}_t$.
In short, its main idea is to weighted-sum all clients' parameters 
according to the influence of their loss values, 
where a larger loss value represents a larger weight for updating $\bar{\theta}^E$.

\subsection{Algorithm Overall Procedure}

The overall procedure of the proposed FedCD framework is summarized in Algorithm \ref{alg1}.
 Firstly, the server randomly initializes the model parameters in the server and  
 distributes these parameters to each client (Lines 1).
After each client takes globally shared parameters are initial values (Line 4),
each client locally trains its local model for $Num$ epochs by minimizing the loss in Eq.(\ref{eq_loss}) (Lines 5-7).
Thirdly, after receiving the parameters of all clients, the server performs global aggregation, parameters $\bar{\theta} ^{E} $
 are updated by applying the aggregation mechanisms in Eq.(\ref{eq_exer}) to collected   $\{\theta_1^E \cdots \theta_T^E\}$ (Lines 9-10). 
The second to third steps are repeated until convergence.

\begin{algorithm}[t]
  
	\caption{Algorithm of FedCD}
	\label{alg1}
	
	
	\begin{algorithmic}[1]
		\STATE $ \{ (\theta ^{S}_{t},\theta ^{E}_{t},\theta ^{I}_{t}) \} ^{T}_{t=1}  \gets$ Randomly initialize $\{ ( {\bar{\theta}} ^{S} , {\bar{\theta}} ^{E} , {\bar{\theta}} ^{I} ) \}$
		
		\FOR{each round $p = 1,2,3,...,P$}

		\FOR{each school index $t \in T $ \textbf{ in parallel}}
		\STATE $\{ \hat{\theta} ^{S}_{t},\hat{\theta} ^{I}_{t} \}  \gets \{ {\bar{\theta}} ^{S},{\bar{\theta}} ^{I} \},\ \theta_t^E \leftarrow \theta_t^E$

  \underline{Client local update:}  
		\FOR{each local round $num = 1,2,3,...,Num $}

		\STATE Optimize Eq.(\ref{eq_loss}) to update $(\theta ^{S}_{t},\theta ^{E}_{t},\theta ^{I}_{t})$
		\ENDFOR
		\ENDFOR
  
		 \underline{Server aggregation:}
		\STATE Update  parameters $ \bar{\theta} ^{E} $ by applying Eq.(\ref{eq_exer}) to $\{\theta_1^E \cdots \theta_T^E\}$
		\ENDFOR

	\end{algorithmic}

\end{algorithm}

\subsection{Privacy Protection Enhanced FedCD}

The risk of privacy leakage can be significantly reduced under the FL framework, since each user retains their private data locally.  
To further prevent potential privacy violations during the upload of model parameters, a local differential privacy strategy is proposed to be integrated into the proposed FedCD. Specifically, we incorporate a zero-mean Laplacian noise into the exercise embedding before uploading it to the server:
\begin{equation}
\label{eq6}
\mathbf{h}_{E}= \mathbf{h}_{E} + \operatorname{Laplacian}(0, \delta) ,
\end{equation}
 $\delta$ represents the noise intensity, and a larger $\delta$  means a stronger capability of privacy protection.
By doing so, the server cannot easily reverse the original data by monitoring the embedding.

\section{EXPERIMENTS}
 This section aims to answer the following research questions:
\begin{itemize}
  
    \item \textbf{RQ1:} How about the effectiveness of the proposed FedCD compared to classical FL approaches under different CDMs?

    \item \textbf{RQ2:} How about the effectiveness of the devised parameter decoupling-based personalization strategy?

    \item \textbf{RQ3:} How about the effectiveness of the proposed  FedCD in solving the client fairness issue?

    \item \textbf{RQ4:} How about the sensitivity of the proposed FedCD to key hyperparameter changes?

    \item \textbf{RQ5:} How about the robustness of the proposed FedCD when integrating local differential privacy?

    \item \textbf{RQ6:} Are there other interesting discussions about FedCD?

\end{itemize}

\subsection{Experimental Setting}

\subsubsection{Datasets}

To implement the FL paradigm, datasets must contain information on students' schools or groupings. Our experiments were conducted on three real-world datasets: ASSIST2009, ASSIST2012, and SLP-Math. Both ASSISTs (short for Assistments) and SLP-Math are public datasets that record students' learning logs from an online education system. During preprocessing, we first removed records with null values and duplicates. We then filtered out students with fewer than 5 records and schools with fewer than 1,000 records. More detailed descriptions and statistics are provided below and in Table \ref{tab1}.
\begin{itemize}

\item The ASSIST2009 dataset captures students' learning logs from the 2009-2010 "Non-skill Builder" period in an online education system. After preprocessing, the dataset comprises 324,659 learning logs from 3,180 students across 21 schools. These logs involve 17,560 exercises and cover 123 distinct concepts.

\item The ASSIST2012 dataset contains students' learning data from the 2012-2013 School Data with Affect. Following preprocessing, this dataset includes 22,470 student logs and 52,447 exercises, spanning 263 distinct concepts.

\item For SLP-Math, we selected the students' math scores, combining the Learner's Personal Info and School Info to form a dataset. The SLP-Math dataset, which includes 10 schools, contains 47,816 records from 1,035 students, with 544 exercises covering 31 distinct concepts.
\end{itemize}



\begin{table}[t]

  \caption{ Statistics of  ASSIST2009, ASSIST2012 and SLP-Math datasets}
  \label{tab1}
      \renewcommand{\arraystretch}{1.1}

  \begin{tabular}{llll}
    \toprule
    Statistics  & ASSIST2009 & ASSIST2012 & SLP-Math\\
    \midrule
    \#of schools & 21 & 141 & 10\\
    \#of records  &  324,659 & 2,570,541 & 47,816\\
    \#of students  & 3,180 & 22,470 & 1,035\\
    \#of exercises  & 17,560 & 52,447 & 544\\
    \#of knowledge concepts  & 123 &  263 & 31\\
  \bottomrule
\end{tabular}
\end{table}





\begin{table*}[t]
	\centering
   \footnotesize
	\caption{Performance of FedCD and all baseline approaches on ASSIST2009. \textbf{Bold:} the best, \underline{Underline:} the runner-up}
        \renewcommand{\arraystretch}{1.4}
	\setlength{\tabcolsep}{0.1mm}
	\begin{tabular}{clrrrrrrrrr}
		\toprule
		\toprule
		\multicolumn{1}{l}{\textbf{Base}} & \textbf{Train/Test ratio} & \multicolumn{3}{c}{\textbf{60\%/40\%}} & \multicolumn{3}{c}{\textbf{70\%/30\%}} & \multicolumn{3}{c}{\textbf{80\%/20\%}} \\
		\cline{2-11}
		\multicolumn{1}{l}{\textbf{Model}} & \textbf{Models} & \multicolumn{1}{l}{\textbf{ACC}} & \multicolumn{1}{l}{\textbf{RMSE}} & \multicolumn{1}{l}{\textbf{AUC}} & \multicolumn{1}{l}{\textbf{ACC}} & \multicolumn{1}{l}{\textbf{RMSE}} & \multicolumn{1}{l}{\textbf{AUC}} & \multicolumn{1}{l}{\textbf{ACC}} & \multicolumn{1}{l}{\textbf{RMSE}} & \multicolumn{1}{l}{\textbf{AUC}} \\
		\hline

		\multirow{9}[0]{*}{\textbf{DINA}} & Base & $0.6557_{\pm0.0006}$ & $0.4768_{\pm0.0004}$ & $0.6899_{\pm0.0014}$ & $0.6591_{\pm0.0005}$ & $0.4755_{\pm0.0002}$ & $0.6937_{\pm0.0013}$ & $0.6602_{\pm0.0011}$ & $0.4749_{\pm0.0004}$ & $0.7048_{\pm0.0012}$ \\
		& FedAvg & $0.6508_{\pm0.0008}$ & $0.4737_{\pm0.0004}$ & $0.6628_{\pm0.0010}$ & $0.6542_{\pm0.0010}$ & $0.4715_{\pm0.0003}$ & $0.6690_{\pm0.0009}$ & $0.6553_{\pm0.0004}$ & $0.4729_{\pm0.0003}$ & $0.6679_{\pm0.0009}$ \\
		& FedAtt & $0.6526_{\pm0.0005}$ & $0.4763_{\pm0.0003}$ & $0.6506_{\pm0.0058}$ & $0.6545_{\pm0.0009}$ & $0.4731_{\pm0.0003}$ & $0.6653_{\pm0.0011}$ & $0.6568_{\pm0.0003}$ & $0.4738_{\pm0.0002}$ & $0.6671_{\pm0.0006}$ \\
		& LoAdaBoost & $0.6359_{\pm0.0008}$ & $0.4795_{\pm0.0003}$ & $0.6642_{\pm0.0019}$ & $0.6360_{\pm0.0028}$ & $0.4795_{\pm0.0010}$ & $0.6651_{\pm0.0033}$ & $0.6345_{\pm0.0029}$ & $0.4802_{\pm0.0009}$ & $0.6680_{\pm0.0027}$ \\
            & F2MF & $0.6518_{\pm0.0008}$ & $0.4759_{\pm0.0002}$ & $0.6499_{\pm0.0011}$ & $0.6545_{\pm0.0007}$ & $0.4715_{\pm0.0001}$ & $0.6680_{\pm0.0005}$ & $0.6549_{\pm0.0008}$ & $0.4729_{\pm0.0002}$ & $0.6680_{\pm0.0013}$ \\
            & PFedRec &$0.6576_{\pm0.0011}$ & $0.4729_{\pm0.0003}$ & $0.6603_{\pm0.0007}$ & $0.6570_{\pm0.0009}$ & $0.4729_{\pm0.0003}$ & $0.6606_{\pm0.0017}$ & $0.6553_{\pm0.0009}$ & $0.4735_{\pm0.0004}$ & $0.6613_{\pm0.0031}$ \\

            & \cellcolor{my_gray}  FedCD & \cellcolor{my_gray}  $\mathbf{0.6768_{\pm0.0004}}$ & \cellcolor{my_gray} $\mathbf{0.4663_{\pm0.0003}}$ & \cellcolor{my_gray} $\mathbf{0.7138_{\pm0.0007}}$ & \cellcolor{my_gray} $\mathbf{0.6781_{\pm0.0009}}$ & \cellcolor{my_gray} $\mathbf{0.4665_{\pm0.0003}}$ & \cellcolor{my_gray} $\mathbf{0.7196_{\pm0.0006}}$ & \cellcolor{my_gray} $\mathbf{0.6830_{\pm0.0013}}$ & \cellcolor{my_gray} $\mathbf{0.4636_{\pm0.0005}}$ & \cellcolor{my_gray} $\mathbf{0.7226_{\pm0.0005}}$ \\
		& \cellcolor{my_gray}  FedCD(FedAvg)& \cellcolor{my_gray} $\underline{0.6728_{\pm0.0011 }}$ & \cellcolor{my_gray} $\underline{0.4673_{\pm0.0004}}$ & \cellcolor{my_gray} $\underline{0.7096_{\pm0.0008}}$ & \cellcolor{my_gray} $\underline{0.6734_{\pm0.0011}}$ & \cellcolor{my_gray} $\underline{0.4676_{\pm0.0003}}$ & \cellcolor{my_gray} $\underline{0.7148_{\pm0.0014}}$ & \cellcolor{my_gray} $\underline{0.6813_{\pm0.0009}}$ & \cellcolor{my_gray} $\underline{0.4643_{\pm0.0003}}$ & \cellcolor{my_gray} $\underline{0.7201_{\pm0.0009}}$ \\
		& \cellcolor{my_gray}  FedCD(FedAtt) & \cellcolor{my_gray} $0.6705_{\pm0.0007}$ & \cellcolor{my_gray} $0.4682_{\pm0.0003}$ & \cellcolor{my_gray} $0.7076_{\pm0.0004}$ & \cellcolor{my_gray} $0.6722_{\pm0.0015}$ & \cellcolor{my_gray} $0.4680_{\pm0.0006}$ & \cellcolor{my_gray} $0.7140_{\pm0.0011}$ & \cellcolor{my_gray} $0.6807_{\pm0.0009}$ & \cellcolor{my_gray} $0.4645_{\pm0.0004}$ & \cellcolor{my_gray} $0.7191_{\pm0.0001}$ \\

		\hline
  
		\multirow{9}[0]{*}{\textbf{NCD}} 
        & Base  & $0.6797_{\pm0.0017}$ & $\underline{0.4494_{\pm0.0002}}$ & $0.7086_{\pm0.0005}$ & $0.6868_{\pm0.0025}$ & $0.4488_{\pm0.0016}$ & $0.7096_{\pm0.0048}$ & $0.6929_{\pm0.0022}$ & $0.4478_{\pm0.0009}$ & $0.7181_{\pm0.0010}$ \\
		& FedAvg & $0.6785_{\pm0.0005}$ & $0.4618_{\pm0.0013}$ & $0.6668_{\pm0.0008}$ & $0.6804_{\pm0.0004}$ & $0.4602_{\pm0.0007}$ & $0.6721_{\pm0.0007}$ & $0.6816_{\pm0.0007}$ & $0.4647_{\pm0.0105}$ & $0.6794_{\pm0.0028}$ \\
		& FedAtt & $0.6744_{\pm0.0014}$ & $0.4699_{\pm0.0011}$ & $0.6624_{\pm0.0012}$ & $0.6752_{\pm0.0017}$ & $0.4691_{\pm0.0005}$ & $0.6660_{\pm0.0010}$ & $0.6755_{\pm0.0024}$ & $0.4706_{\pm0.0063}$ & $0.6722_{\pm0.0019}$ \\
		& LoAdaBoost & $0.6935_{\pm0.0007}$ & $0.4511_{\pm0.0003}$ & $0.6982_{\pm0.0004}$ & $0.6938_{\pm0.0019}$ & $0.4505_{\pm0.0007}$ & $0.6998_{\pm0.0006}$ & $0.6978_{\pm0.0005}$ & $0.4502_{\pm0.0003}$ & $0.7045_{\pm0.0006}$ \\
            & F2MF & $0.6781_{\pm0.0003}$ & $0.4619_{\pm0.0015}$ & $0.6668_{\pm0.0012}$ & $0.6806_{\pm0.0007}$ & $0.4595_{\pm0.0007}$ & $0.6714_{\pm0.0011}$ & $0.6817_{\pm0.0004}$ & $0.4597_{\pm0.0003}$ & $0.6778_{\pm0.0004}$ \\
            & PFedRec &$0.6831_{\pm0.0005}$ & $0.4614_{\pm0.0066}$ & $0.6996_{\pm0.0075}$ & $0.6927_{\pm0.0006}$ & $0.4608_{\pm0.0013}$ & $0.7080_{\pm0.0007}$ & $0.6986_{\pm0.0004}$ & $0.4604_{\pm0.0003}$ & $0.7170_{\pm0.0002}$ \\
            
            & \cellcolor{my_gray}  FedCD & \cellcolor{my_gray} $\mathbf{0.7017_{\pm0.0047}}$ & \cellcolor{my_gray} $\mathbf{0.4483_{\pm0.0047}}$ & \cellcolor{my_gray} $\mathbf{0.7320_{\pm0.0011}}$ & \cellcolor{my_gray} $\mathbf{0.7114_{\pm0.0019}}$ & \cellcolor{my_gray} $0.4487_{\pm0.0097}$ & \cellcolor{my_gray} $\mathbf{0.7389_{\pm0.0031}}$ & \cellcolor{my_gray} $\mathbf{0.7201_{\pm0.0013}}$ & \cellcolor{my_gray} $0.4539_{\pm0.0043}$ & \cellcolor{my_gray} $0.7413_{\pm0.0058}$ \\
		& \cellcolor{my_gray}  FedCD(FedAvg) & \cellcolor{my_gray} $\underline{0.7013_{\pm0.0016}}$ & \cellcolor{my_gray} $0.4502_{\pm0.0025}$ & \cellcolor{my_gray} $\underline{0.7257_{\pm0.0016}}$ & \cellcolor{my_gray} $\underline{0.7054_{\pm0.0026}}$ & \cellcolor{my_gray} $\mathbf{0.4452_{\pm0.0008}}$ & \cellcolor{my_gray} $\underline{0.7325_{\pm0.0011}}$ & \cellcolor{my_gray} $0.7097_{\pm0.0018}$ & \cellcolor{my_gray} $\mathbf{0.4423_{\pm0.0013}}$ & \cellcolor{my_gray} $\mathbf{0.7430_{\pm0.0032}}$ \\
		& \cellcolor{my_gray}  FedCD(FedAtt) & \cellcolor{my_gray} $0.7012_{\pm0.0020}$ & \cellcolor{my_gray} $0.4515_{\pm0.0025}$ & \cellcolor{my_gray} $0.7245_{\pm0.0021}$ & \cellcolor{my_gray} $0.7020_{\pm0.0018}$ & \cellcolor{my_gray} $\underline{0.4473_{\pm0.0016}}$ & \cellcolor{my_gray} $0.7319_{\pm0.0026}$ & \cellcolor{my_gray} $\underline{0.7118_{\pm0.0022}}$ & \cellcolor{my_gray} $\underline{0.4435_{\pm0.0033}}$ & \cellcolor{my_gray} $\underline{0.7426_{\pm0.0020}}$ \\
  
		\hline
  
		\multirow{9}[0]{*}{\textbf{KaNCD}} 
        & Base & $0.6975_{\pm0.0123}$ & $\mathbf{0.4621_{\pm0.0130}}$ & $\mathbf{0.7376_{\pm0.0034}}$ & $0.7024_{\pm0.0042}$ & $\mathbf{0.4605_{\pm0.0109}}$ & $\mathbf{0.7397_{\pm0.0034}}$ & $0.7013_{\pm0.0051}$ & $\mathbf{0.4596_{\pm0.0041}}$ & $\mathbf{0.7420_{\pm0.0013}}$ \\
		& FedAvg & $0.6734_{\pm0.0042}$ & $0.5475_{\pm0.0187}$ & $0.6750_{\pm0.0086}$ & $0.6727_{\pm0.0035}$ & $0.5552_{\pm0.0129}$ & $0.6752_{\pm0.0047}$ & $0.6815_{\pm0.0054}$ & $0.5434_{\pm0.0171}$ & $0.6799_{\pm0.0063}$ \\
		& FedAtt & $0.6747_{\pm0.0056}$ & $0.5457_{\pm0.0114}$ & $0.6728_{\pm0.0067}$ & $0.6797_{\pm0.0044}$ & $0.5493_{\pm0.0134}$ & $0.6700_{\pm0.0113}$ & $0.6773_{\pm0.0008}$ & $0.5561_{\pm0.0113}$ & $0.6688_{\pm0.0079}$ \\
		& LoAdaBoost & $0.6643_{\pm0.0019}$ & $0.5711_{\pm0.0019}$ & $0.7162_{\pm0.0015}$ & $0.6664_{\pm0.0042}$ & $0.5690_{\pm0.0032}$ & $0.7198_{\pm0.0011}$ & $0.6717_{\pm0.0030}$ & $0.5439_{\pm0.0474}$ & $0.7099_{\pm0.0256}$ \\
            & F2MF & $0.6752_{\pm0.0053}$ & $0.5456_{\pm0.0229}$ & $0.6732_{\pm0.0042}$ & $0.6680_{\pm0.0029}$ & $0.5717_{\pm0.0033}$ & $0.6685_{\pm0.0020}$ & $0.6785_{\pm0.0059}$ & $0.5550_{\pm0.0167}$ & $0.6776_{\pm0.0038}$ \\
            & PFedRec &$0.6772_{\pm0.0036}$ & $0.5006_{\pm0.0021}$ & $0.6906_{\pm0.0004}$ & $0.6895_{\pm0.0007}$ & $0.5056_{\pm0.0016}$ & $0.6968_{\pm0.0006}$ & $0.6895_{\pm0.0011}$ & $0.5000_{\pm0.0151}$ & $0.6994_{\pm0.0075}$ \\

            & \cellcolor{my_gray}  FedCD & \cellcolor{my_gray} $\mathbf{0.7105_{\pm0.0013}}$ & \cellcolor{my_gray} $0.5360_{\pm0.0011}$ & \cellcolor{my_gray} $0.7044_{\pm0.0021}$ & \cellcolor{my_gray} $\mathbf{0.7154_{\pm0.0029}}$ & \cellcolor{my_gray} $\underline{0.5317_{\pm0.0027}}$ & \cellcolor{my_gray} $0.7095_{\pm0.0020}$ & \cellcolor{my_gray} $\mathbf{0.7196_{\pm0.0022}}$ & \cellcolor{my_gray} $\underline{0.5277_{\pm0.0022}}$ & \cellcolor{my_gray} $0.7107_{\pm0.0021}$ \\
		& \cellcolor{my_gray}  FedCD(FedAvg) & \cellcolor{my_gray} $\underline{0.7067_{\pm0.0004}}$ & \cellcolor{my_gray} $0.5363_{\pm0.0015}$ & \cellcolor{my_gray} $\underline{0.7169_{\pm0.0019}}$ & \cellcolor{my_gray} $\underline{0.7119_{\pm0.0023}}$ & \cellcolor{my_gray} $0.5319_{\pm0.0025}$ & \cellcolor{my_gray} $0.7201_{\pm0.0022}$ & \cellcolor{my_gray} $\underline{0.7135_{\pm0.0022}}$ & \cellcolor{my_gray} $0.5306_{\pm0.0017}$ & \cellcolor{my_gray} $\underline{0.7222_{\pm0.0010}}$ \\
		& \cellcolor{my_gray}  FedCD(FedAtt) & \cellcolor{my_gray} $0.7065_{\pm0.0020}$ & \cellcolor{my_gray} $\underline{0.5352_{\pm0.0028}}$ & \cellcolor{my_gray} $0.7158_{\pm0.0023}$ & \cellcolor{my_gray} $0.7108_{\pm0.0013}$ & \cellcolor{my_gray} $0.5327_{\pm0.0018}$ & \cellcolor{my_gray} $\underline{0.7201_{\pm0.0010}}$ & \cellcolor{my_gray} $0.7108_{\pm0.0022}$ & \cellcolor{my_gray} $0.5325_{\pm0.0028}$ & \cellcolor{my_gray} $0.7201_{\pm0.0023}$ \\

        \hline

           \textbf{GUM}
        & HPFL & $0.6740_{\pm0.0004}$ & $0.4661_{\pm0.0011}$ & $0.6729_{\pm0.0004}$ & $0.6772_{\pm0.0013}$ & $0.5473_{\pm0.0143}$ & $0.6603_{\pm0.0115}$ & $0.6826_{\pm0.0003}$ & $0.5514_{\pm0.0003}$ & $0.6612_{\pm0.0015}$ \\
        
        \hline
		
		      \toprule
		\toprule
	\end{tabular}%

    \begin{tabular}{l|lr|lr|lr|lr|lr|lr}
    \toprule
    \textbf{Number of times FedCD wins X} & vs. base \quad &  20 \quad & vs. FedAvg \quad & 27 \quad & vs. FedAtt \quad & 27 \quad & vs. LoAdaBoost \quad & 24 \quad & vs. F2MF \quad & 27 \quad & vs. PFedRec \quad & 24 \quad \\
    \bottomrule
    \end{tabular}%

	\label{tab2}%
\end{table*}%

\begin{table*}[t]
	\centering
   \footnotesize
	\caption{Performance of FedCD and all baseline approaches on ASSIST2012. \textbf{Bold:} the best, \underline{Underline:} the runner-up}
    \renewcommand{\arraystretch}{1.4}
	\setlength{\tabcolsep}{0.1mm}
	\begin{tabular}{clrrrrrrrrr}
		\toprule
		\toprule
		\multicolumn{1}{l}{\textbf{Base}} & \textbf{Train/Test ratio} & \multicolumn{3}{c}{\textbf{60\%/40\%}} & \multicolumn{3}{c}{\textbf{70\%/30\%}} & \multicolumn{3}{c}{\textbf{80\%/20\%}} \\
		\cline{2-11}
		\multicolumn{1}{l}{\textbf{Model}} & \textbf{Models} & \multicolumn{1}{l}{\textbf{ACC}} & \multicolumn{1}{l}{\textbf{RMSE}} & \multicolumn{1}{l}{\textbf{AUC}} & \multicolumn{1}{l}{\textbf{ACC}} & \multicolumn{1}{l}{\textbf{RMSE}} & \multicolumn{1}{l}{\textbf{AUC}} & \multicolumn{1}{l}{\textbf{ACC}} & \multicolumn{1}{l}{\textbf{RMSE}} & \multicolumn{1}{l}{\textbf{AUC}} \\
		\hline

		\multirow{9}[0]{*}{\textbf{DINA}} 
        & Base  & $\mathbf{0.6924_{\pm0.0006}}$ & $\mathbf{0.4603_{\pm0.0002}}$ & $0.6522_{\pm0.0004}$ & $\mathbf{0.6923_{\pm0.0007}}$ & $\mathbf{0.4607_{\pm0.0004}}$ & $0.6511_{\pm0.0011}$ & $\mathbf{0.6929_{\pm0.0006}}$ & $\mathbf{0.4605_{\pm0.0003}}$ & $0.6523_{\pm0.0007}$ \\
		& FedAvg & $0.6574_{\pm0.0009}$ & $0.4727_{\pm0.0003}$ & $0.6328_{\pm0.0005}$ & $0.6580_{\pm0.0006}$ & $0.4727_{\pm0.0001}$ & $0.6332_{\pm0.0008}$ & $0.6592_{\pm0.0009}$ & $0.4721_{\pm0.0002}$ & $0.6355_{\pm0.0017}$ \\
		& FedAtt & $0.6577_{\pm0.0005}$ & $0.4726_{\pm0.0003}$ & $0.6323_{\pm0.0009}$ & $0.6592_{\pm0.0006}$ & $0.4723_{\pm0.0003}$ & $0.6344_{\pm0.0013}$ & $0.6595_{\pm0.0007}$ & $0.4721_{\pm0.0003}$ & $0.6354_{\pm0.0017}$ \\
		& LoAdaBoost & $0.6524_{\pm0.0007}$ & $0.4736_{\pm0.0003}$ & $0.6305_{\pm0.0007}$ & $0.6527_{\pm0.0011}$ & $0.4736_{\pm0.0004}$ & $0.6324_{\pm0.0007}$ & $0.6547_{\pm0.0010}$ & $0.4729_{\pm0.0003}$ & $0.6349_{\pm0.0007}$ \\
            & F2MF & $0.6574_{\pm0.0007}$ & $0.4725_{\pm0.0003}$ & $0.6333_{\pm0.0005}$ & $0.6583_{\pm0.0006}$ & $0.4726_{\pm0.0002}$ & $0.6334_{\pm0.0011}$ & $0.6594_{\pm0.0009}$ & $0.4720_{\pm0.0004}$ & $0.6356_{\pm0.0011}$ \\
            & PFedRec &$0.6869_{\pm0.0002}$ & $0.4614_{\pm0.0000}$ & $0.6472_{\pm0.0015}$ & $0.6872_{\pm0.0002}$ & $0.4614_{\pm0.0002}$ & $0.6494_{\pm0.0025}$ & $0.6877_{\pm0.0005}$ & $0.4612_{\pm0.0003}$ & $0.6490_{\pm0.0020}$ \\
        
            & \cellcolor{my_gray}  FedCD & \cellcolor{my_gray} $0.6823_{\pm0.0012}$ & \cellcolor{my_gray} $0.4636_{\pm0.0005}$ & \cellcolor{my_gray} $0.6617_{\pm0.0016}$ & \cellcolor{my_gray} $0.6831_{\pm0.0004}$ & \cellcolor{my_gray} $0.4636_{\pm0.0002}$ & \cellcolor{my_gray} $0.6549_{\pm0.0032}$ & \cellcolor{my_gray} $0.6837_{\pm0.0008}$ & \cellcolor{my_gray} $0.4630_{\pm0.0003}$ & \cellcolor{my_gray} $0.6733_{\pm0.0066}$ \\
		& \cellcolor{my_gray}  FedCD(FedAvg) & \cellcolor{my_gray} $\underline{0.6861_{\pm0.0006}}$ & \cellcolor{my_gray} $\underline{0.4612_{\pm0.0003}}$ & \cellcolor{my_gray} $\underline{0.6684_{\pm0.0030}}$ & \cellcolor{my_gray} $0.6865_{\pm0.0003}$ & \cellcolor{my_gray} $0.4611_{\pm0.0002}$ & \cellcolor{my_gray} $\underline{0.6723_{\pm0.0058}}$ & \cellcolor{my_gray} $\underline{0.6882_{\pm0.0004}}$ & \cellcolor{my_gray} $\underline{0.4606_{\pm0.0001}}$ & \cellcolor{my_gray} $\mathbf{0.6784_{\pm0.0036}}$ \\
		& \cellcolor{my_gray}  FedCD(FedAtt) & \cellcolor{my_gray} $0.6855_{\pm0.0006}$ & \cellcolor{my_gray} $0.4613_{\pm0.0002}$ & \cellcolor{my_gray} $\mathbf{0.6705_{\pm0.0038}}$ & \cellcolor{my_gray} $\underline{0.6868_{\pm0.0002}}$ & \cellcolor{my_gray} $\underline{0.4609_{\pm0.0000}}$ & \cellcolor{my_gray} $\mathbf{0.6742_{\pm0.0020}}$ & \cellcolor{my_gray} $0.6878_{\pm0.0002}$ & \cellcolor{my_gray} $0.4607_{\pm0.0002}$ & \cellcolor{my_gray} $\underline{0.6777_{\pm0.0026}}$ \\
  
		\hline
  
		\multirow{9}[0]{*}{\textbf{NCD}} 
        & Base  & $0.7175_{\pm0.0003}$ & $\mathbf{0.4360_{\pm0.0002}}$ & $\mathbf{0.7038_{\pm0.0006}}$ & $0.7181_{\pm0.0002}$ & $\mathbf{0.4343_{\pm0.0012}}$ & $\mathbf{0.7102_{\pm0.0071}}$ & $0.7192_{\pm0.0002}$ & $\mathbf{0.4341_{\pm0.0002}}$ & $\mathbf{0.7094_{\pm0.0003}}$ \\
		& FedAvg & $0.7086_{\pm0.0003}$ & $0.4456_{\pm0.0002}$ & $0.6688_{\pm0.0002}$ & $0.7117_{\pm0.0002}$ & $0.4474_{\pm0.0002}$ & $0.6721_{\pm0.0003}$ & $0.7122_{\pm0.0008}$ & $0.4492_{\pm0.0004}$ & $0.6737_{\pm0.0001}$ \\
		& FedAtt & $0.7095_{\pm0.0002}$ & $0.4464_{\pm0.0002}$ & $0.6680_{\pm0.0002}$ & $0.7117_{\pm0.0002}$ & $0.4483_{\pm0.0004}$ & $0.6718_{\pm0.0002}$ & $0.7096_{\pm0.0029}$ & $0.4488_{\pm0.0022}$ & $0.6730_{\pm0.0007}$ \\
		& LoAdaBoost & $0.7113_{\pm0.0004}$ & $0.4545_{\pm0.0003}$ & $0.6645_{\pm0.0001}$ & $0.7135_{\pm0.0003}$ & $0.4540_{\pm0.0005}$ & $0.6659_{\pm0.0002}$ & $0.7100_{\pm0.0006}$ & $0.4532_{\pm0.0012}$ & $0.6663_{\pm0.0003}$ \\
            & F2MF & $0.7118_{\pm0.0002}$ & $0.4481_{\pm0.0005}$ & $0.6701_{\pm0.0001}$ & $0.7102_{\pm0.0002}$ & $0.4457_{\pm0.0002}$ & $0.6725_{\pm0.0000}$ & $0.7131_{\pm0.0004}$ & $0.4467_{\pm0.0007}$ & $0.6741_{\pm0.0003}$ \\
            & PFedRec &$0.7205_{\pm0.0001}$ & $0.4363_{\pm0.0001}$ & $0.6965_{\pm0.0000}$ & $0.7214_{\pm0.0003}$ & $0.4357_{\pm0.0002}$ & $0.7008_{\pm0.0000}$ & $0.7224_{\pm0.0001}$ & $0.4347_{\pm0.0001}$ & $0.7045_{\pm0.0000}$ \\
            
            & \cellcolor{my_gray}  FedCD & \cellcolor{my_gray} $0.7177_{\pm0.0003}$ & \cellcolor{my_gray} $0.4491_{\pm0.0004}$ & \cellcolor{my_gray} $0.7014_{\pm0.0005}$ & \cellcolor{my_gray} $0.7181_{\pm0.0007}$ & \cellcolor{my_gray} $0.4471_{\pm0.0009}$ & \cellcolor{my_gray} $0.7045_{\pm0.0004}$ & \cellcolor{my_gray} $0.7221_{\pm0.0003}$ & \cellcolor{my_gray} $0.4439_{\pm0.0013}$ & \cellcolor{my_gray} $\underline{0.7092_{\pm0.0016}}$ \\
		& \cellcolor{my_gray}  FedCD(FedAvg) & \cellcolor{my_gray} $\mathbf{0.7227_{\pm0.0003}}$ & \cellcolor{my_gray} $\underline{0.4370_{\pm0.0003}}$ & \cellcolor{my_gray} $0.7022_{\pm0.0006}$ & \cellcolor{my_gray} $\underline{0.7240_{\pm0.0004}}$ & \cellcolor{my_gray} $\underline{0.4360_{\pm0.0002}}$ & \cellcolor{my_gray} $0.7056_{\pm0.0005}$ & \cellcolor{my_gray} $\mathbf{0.7260_{\pm0.0003}}$ & \cellcolor{my_gray} $0.4352_{\pm0.0003}$ & \cellcolor{my_gray} $0.7081_{\pm0.0004}$ \\
		& \cellcolor{my_gray}  FedCD(FedAtt) & \cellcolor{my_gray} $\underline{0.7225_{\pm0.0003}}$ & \cellcolor{my_gray} $0.4370_{\pm0.0002}$ & \cellcolor{my_gray} $\underline{0.7023_{\pm0.0008}}$ & \cellcolor{my_gray} $\mathbf{0.7246_{\pm0.0005}}$ & \cellcolor{my_gray} $0.4362_{\pm0.0002}$ & \cellcolor{my_gray} $\underline{0.7056_{\pm0.0010}}$ & \cellcolor{my_gray} $\underline{0.7258_{\pm0.0003}}$ & \cellcolor{my_gray} $\underline{0.4350_{\pm0.0002}}$ & \cellcolor{my_gray} $0.7082_{\pm0.0006}$ \\

		\hline
  
		\multirow{9}[0]{*}{\textbf{KaNCD}} 
        & Base & $\mathbf{0.7163_{\pm0.0030}}$ & $\mathbf{0.4473_{\pm0.0154}}$ & $\mathbf{0.7158_{\pm0.0026}}$ & $\mathbf{0.7125_{\pm0.0017}}$ & $\mathbf{0.4506_{\pm0.0174}}$ & $\mathbf{0.7134_{\pm0.0037}}$ & $\mathbf{0.7219_{\pm0.0022}}$ & $\mathbf{0.4461_{\pm0.0099}}$ & $\mathbf{0.7182_{\pm0.0033}}$ \\
		& FedAvg & $0.6807_{\pm0.0195}$ & $0.5083_{\pm0.0312}$ & $0.6494_{\pm0.0061}$ & $0.6693_{\pm0.0192}$ & $0.5152_{\pm0.0338}$ & $0.6534_{\pm0.0041}$ & $0.6616_{\pm0.0143}$ & $0.5252_{\pm0.0323}$ & $0.6515_{\pm0.0069}$ \\
		& FedAtt & $0.6874_{\pm0.0175}$ & $\underline{0.4976_{\pm0.0056}}$ & $0.6474_{\pm0.0039}$ & $0.6591_{\pm0.0056}$ & $0.5552_{\pm0.0131}$ & $0.6593_{\pm0.0020}$ & $0.6625_{\pm0.0141}$ & $0.5451_{\pm0.0286}$ & $0.6573_{\pm0.0028}$ \\
		& LoAdaBoost & $0.6728_{\pm0.0075}$ & $0.5171_{\pm0.0119}$ & $0.6581_{\pm0.0015}$ & $0.6747_{\pm0.0216}$ & $\underline{0.5059_{\pm0.0262}}$ & $0.6527_{\pm0.0021}$ & $0.6729_{\pm0.0166}$ & $\underline{0.5191_{\pm0.0177}}$ & $0.6555_{\pm0.0040}$ \\
            & F2MF &$0.6597_{\pm0.0206}$ & $0.5515_{\pm0.0448}$ & $0.6573_{\pm0.0056}$ & $0.6739_{\pm0.0240}$ & $0.5273_{\pm0.0483}$ & $0.6561_{\pm0.0068}$ & $0.6734_{\pm0.0208}$ & $0.5252_{\pm0.0360}$ & $0.6547_{\pm0.0053}$ \\
            & PFedRec &$0.6799_{\pm0.0017}$ & $0.5444_{\pm0.0359}$ & $0.6534_{\pm0.0195}$ & $0.6950_{\pm0.0065}$ & $0.4864_{\pm0.0011}$ & $0.6860_{\pm0.0002}$ & $0.6768_{\pm0.0065}$ & $0.5348_{\pm0.0425}$ & $0.6630_{\pm0.0306}$ \\

            & \cellcolor{my_gray}  FedCD & \cellcolor{my_gray} $\underline{0.7085_{\pm0.0049}}$ & \cellcolor{my_gray} $0.5350_{\pm0.0014}$ & \cellcolor{my_gray} $0.6631_{\pm0.0142}$ & \cellcolor{my_gray} $\underline{0.7102_{\pm0.0026}}$ & \cellcolor{my_gray} $0.5351_{\pm0.0022}$ & \cellcolor{my_gray} $0.6596_{\pm0.0040}$ & \cellcolor{my_gray} $\underline{0.7121_{\pm0.0016}}$ & \cellcolor{my_gray} $0.5335_{\pm0.0013}$ & \cellcolor{my_gray} $0.6639_{\pm0.0022}$ \\
		& \cellcolor{my_gray}  FedCD(FedAvg) & \cellcolor{my_gray} $0.7006_{\pm0.0017}$ & \cellcolor{my_gray} $0.5352_{\pm0.0016}$ & \cellcolor{my_gray} $0.6849_{\pm0.0010}$ & \cellcolor{my_gray} $0.7015_{\pm0.0011}$ & \cellcolor{my_gray} $0.5357_{\pm0.0036}$ & \cellcolor{my_gray} $\underline{0.6841_{\pm0.0014}}$ & \cellcolor{my_gray} $0.7038_{\pm0.0009}$ & \cellcolor{my_gray} $0.5321_{\pm0.0008}$ & \cellcolor{my_gray} $\underline{0.6865_{\pm0.0017}}$ \\
		& \cellcolor{my_gray}  FedCD(FedAtt) & \cellcolor{my_gray} $0.7025_{\pm0.0017}$ & \cellcolor{my_gray} $0.5336_{\pm0.0018}$ & \cellcolor{my_gray} $\underline{0.6853_{\pm0.0005}}$ & \cellcolor{my_gray} $0.7021_{\pm0.0020}$ & \cellcolor{my_gray} $0.5353_{\pm0.0041}$ & \cellcolor{my_gray} $0.6836_{\pm0.0026}$ & \cellcolor{my_gray} $0.7045_{\pm0.0021}$ & \cellcolor{my_gray} $0.5317_{\pm0.0016}$ & \cellcolor{my_gray} $0.6861_{\pm0.0012}$ \\

        \hline
        
        \textbf{GUM}
        & HPFL & $0.7023_{\pm0.0001}$ & $0.4518_{\pm0.0033}$ & $0.6556_{\pm0.0014}$ & $0.7013_{\pm0.0002}$ & $0.4514_{\pm0.0003}$ & $0.6578_{\pm0.0003}$ & $0.7019_{\pm0.0003}$ & $0.4643_{\pm0.0226}$ & $0.6619_{\pm0.0057}$ \\

        \hline
		
				\toprule
		\toprule
	\end{tabular}%

    \begin{tabular}{l|lr|lr|lr|lr|lr|lr}
    \toprule
    \textbf{Number of times FedCD wins X} & vs. base \quad &  6 \quad & vs. FedAvg \quad & 23 \quad & vs. FedAtt \quad & 25 \quad & vs. LoAdaBoost \quad & 24 \quad & vs. F2MF \quad & 23 \quad & vs. PFedRec \quad & 13 \quad \\
    \bottomrule
    \end{tabular}%
        
	\label{tab3}%
\end{table*}%

\subsubsection{Baseline Approaches} 
\label{sec:variants}

Three CDMs and five FL approaches were employed to demonstrate the effectiveness of the proposed FedCD framework. Specifically, three representative CDMs were utilized as the base models under the FL paradigm to showcase the scalability of FedCD, including one classical model, DINA~\cite{Torre2009DINA}, and two state-of-the-art (SOTA) models, NCD~\cite{wang2020neural} and KaNCD~\cite{wang2022kancd}. Two classical FL approaches, FedAvg~\cite{fedavg} and FedAtt~\cite{fedatt}, one SOTA FL approach, LoAdaBoost~\cite{huang2020loadaboost}, one personalization FL recommendation approach, PFedRec~\cite{zhang2023dual}, and one fairness-aware FL recommendation approach, F2MF~\cite{Liu2022Fairness}, were compared to highlight the superiority of the proposed FedCD. Additionally, the FL approach for CD, named HPFL~\cite{wu2021federated}, was also compared to demonstrate the effectiveness of FedCD in addressing the fairness issue. More details about the aforementioned CDMs and FL approaches can be found as follows.


\begin{itemize}
\item \textbf{DINA} models students' mastery as a multidimensional knowledge concept vector and considers the factors of students' guessing and slipping.
\item \textbf{NCD} is one of the latest CDMs that introduces deep learning. It employs a neural network instead of a manually designed prediction function to predict students' responses.
\item \textbf{KaNCD} explores uncovered knowledge concepts of a student by extending NeuralCD while considering knowledge associations.
\end{itemize}

The FL approaches compared in this paper can be categorized into three groups. 
The first group comprises general federated learning approaches, 
such as FedAvg, FedAtt, and LoAdaBoost.
The second group includes one FL approach based on personalization techniques, i.e., PFedRec, 
and one FL approach designed to address fairness issues, i.e., F2MF.
The final group consists of the proposed FedCD and its variants. 
Additionally, there is an FL approach for CD, termed HPFL, which is the first work to apply FL techniques to CD.
Detailed descriptions of these approaches are provided below.
\begin{itemize}
\item \textbf{FedAvg:} This is a classical federated learning algorithm that averages local model updates from participating clients to create a global model. It is widely used due to its simplicity and scalability across various applications.
\item \textbf{FedAtt:} This approach extends FedAvg by incorporating an attention mechanism to assign adaptive weights to client updates based on their importance, thereby enhancing the robustness and fairness of the global model.
\item \textbf{LoAdaBoost:} This is a federated boosting approach that iteratively adjusts the weights of clients, emphasizing those with lower performance to achieve better accuracy and fairness in the global model.
\item \textbf{Fairness-aware Federated Matrix Factorization (F2MF):} This approach aims to ensure fairness across different user groups in federated learning. To achieve this, it facilitates the communication of group-level statistics during the federated optimization process. Additionally, it incorporates differential privacy techniques to protect users' group information, particularly in scenarios where privacy protection is essential.
\item \textbf{Personalized Federated Recommendation (PFedRec):} This approach introduces a personalized federated learning framework tailored for recommendation tasks, combining global knowledge aggregation with personalized model adaptation to enhance recommendation accuracy for individual users.
\item \textbf{Hierarchical Personalized Federated Learning (HPFL):} This approach designs a local user model GUM that applies different aggregation strategies for private and public components. Despite variations in settings across different clients, the model ensures both similarity and personalization in its components.
\end{itemize}
The descriptions of the variants of FedCD have been presented previously.

There are several variants of FedCD designed to validate the effectiveness of its different components, described as follows:
\begin{itemize}
  
    \item \textit{FedCD(FedAvg)}: Similar to FedCD, but uses average aggregation from FedAvg for the exercise embedding parameters.
    \item \textit{FedCD(FedAvg)-s.}: Same as \textit{FedCD(FedAvg)}, but also applies average aggregation for the student embedding parameters. 
    \item \textit{FedCD(FedAvg)-p.d.p.}: 
    FedCD without the parameter decoupling-based personalization strategy, 
    i.e., sharing the diagnostic function-related parameters $\theta_t^I$ across all clients, and using average aggregation for both exercise embedding parameters and diagnostic function-related parameters.
    
    \item \textit{FedCD(FedAtt)}: Similar to FedCD, but employs attention-based aggregation from FedAtt for the exercise embedding parameters.
    \item \textit{FedCD(FedAtt)-s.}: Same as \textit{FedCD(FedAtt)}, but also applies attention-based aggregation for the student embedding parameters. 
    \item \textit{FedCD(FedAtt)-p.d.p.}: Same as \textit{FedCD(FedAvg)-p.d.p.}, but uses attention-based aggregation for the exercise embedding parameters and the diagnostic function parameters.
\end{itemize}

\subsubsection{Evaluation Metrics}
According to the experiences of previous research~\cite{wang2020neural,gao2021graph,cui2025rebalancing}, 
three metrics were adopted to measure the performance of  all CDMs~\cite{yang2023designing}, including \textit{area under the curve }(AUC), \textit{accuracy} (ACC), and \textit{root mean square error} (RMSE).

\subsubsection{Parameter Settings}
For the proposed FedCD, 
its dimension $D$ was set to the number of concepts, i.e., $K$,
$\gamma$ was set  to 0.1, 
the maximal number of rounds $P$ was set to 100, 
the number of local training epochs $Num$ was set to 5, 
the Adam optimizer was employed for local model training with a learning rate of 0.001 and batch size of 128.
For comprehensive comparisons, three splitting settings ( 60\%/40\%, 70\%/30\%, and 80\%/20\%) were adopted to get training and testing datasets. 
All compared CDMs followed the settings in their original papers, 
and FedAvg, FedAtt, and LoAdaBoost hold the same settings as FedCD. 
All models were implemented by PyTorch, and all experiments were conducted on an NVIDIA 3090 GPU.
The source code of FedCD is at \url{https://anonymous.4open.science/r/FedCD-67F8/FedCD-main.zip}.

\subsection{Overall Performance Comparison (RQ1)}
\label{main_exp}

\begin{table*}[!t]
	\centering
   \footnotesize
	\caption{Performance of FedCD and all baseline approaches on SLP-Math. \textbf{Bold:} the best, \underline{Underline:} the runner-up}
        \renewcommand{\arraystretch}{1.4}
	\setlength{\tabcolsep}{0.1mm}
	\begin{tabular}{clrrrrrrrrr}
		\toprule
		\toprule
		\multicolumn{1}{l}{\textbf{Base}} & \textbf{Train/Test ratio} & \multicolumn{3}{c}{\textbf{60\%/40\%}} & \multicolumn{3}{c}{\textbf{70\%/30\%}} & \multicolumn{3}{c}{\textbf{80\%/20\%}} \\
		\hline
		\multicolumn{1}{l}{\textbf{Model}} & \textbf{Models} & \multicolumn{1}{l}{\textbf{ACC}} & \multicolumn{1}{l}{\textbf{RMSE}} & \multicolumn{1}{l}{\textbf{AUC}} & \multicolumn{1}{l}{\textbf{ACC}} & \multicolumn{1}{l}{\textbf{RMSE}} & \multicolumn{1}{l}{\textbf{AUC}} & \multicolumn{1}{l}{\textbf{ACC}} & \multicolumn{1}{l}{\textbf{RMSE}} & \multicolumn{1}{l}{\textbf{AUC}} \\
		\hline
  
		\multirow{9}[0]{*}{\textbf{DINA}} & Base & $0.6470_{\pm0.0010}$ & $0.4781_{\pm0.0004}$ & $0.7011_{\pm0.0045}$ & $0.6498_{\pm0.0014}$ & $0.4766_{\pm0.0003}$ & $0.7071_{\pm0.0020}$ & $0.6486_{\pm0.0009}$ & $0.4776_{\pm0.0001}$ & $0.7043_{\pm0.0023}$ \\

		& FedAvg & $0.6354_{\pm0.0014}$ & $0.4783_{\pm0.0005}$ & $0.6962_{\pm0.0028}$ & $0.6388_{\pm0.0026}$ & $0.4777_{\pm0.0007}$ & $0.7022_{\pm0.0019}$ & $0.6427_{\pm0.0024}$ & $0.4780_{\pm0.0005}$ & $0.7059_{\pm0.0019}$ \\
		& FedAtt & $0.6446_{\pm0.0015}$ & $0.4759_{\pm0.0006}$ & $0.6974_{\pm0.0018}$ & $0.6472_{\pm0.0021}$ & $0.4745_{\pm0.0006}$ & $0.7028_{\pm0.0022}$ & $0.6502_{\pm0.0009}$ & $0.4743_{\pm0.0005}$ & $0.7068_{\pm0.0033}$ \\
		& LoAdaBoost & $0.6412_{\pm0.0012}$ & $0.4771_{\pm0.0006}$ & $0.7039_{\pm0.0040}$ & $0.6348_{\pm0.0026}$ & $0.4785_{\pm0.0003}$ & $0.7055_{\pm0.0051}$ & $0.6418_{\pm0.0036}$ & $0.4784_{\pm0.0004}$ & $0.7135_{\pm0.0035}$ \\
            & F2MF & $0.6381_{\pm0.0015}$ & $0.4781_{\pm0.0004}$ & $0.6964_{\pm0.0024}$ & $0.6364_{\pm0.0014}$ & $0.4786_{\pm0.0005}$ & $0.6969_{\pm0.0013}$ & $0.6446_{\pm0.0010}$ & $0.4774_{\pm0.0003}$ & $0.7071_{\pm0.0023}$ \\
            & PFedRec &$0.6399_{\pm0.0016}$ & $0.4794_{\pm0.0006}$ & $0.7004_{\pm0.0073}$ & $0.6382_{\pm0.0030}$ & $0.4811_{\pm0.0011}$ & $0.7100_{\pm0.0088}$ & $0.6388_{\pm0.0023}$ & $0.4803_{\pm0.0008}$ & $0.7008_{\pm0.0083}$ \\

            & \cellcolor{my_gray}  FedCD & \cellcolor{my_gray} $\mathbf{0.6800_{\pm0.0012}}$ & \cellcolor{my_gray} $0.4664_{\pm0.0004}$ & \cellcolor{my_gray} $\mathbf{0.7661_{\pm0.0015}}$ & \cellcolor{my_gray} $\mathbf{0.6868_{\pm0.0007}}$ & \cellcolor{my_gray} $\mathbf{0.4624_{\pm0.0006}}$ & \cellcolor{my_gray} $\mathbf{0.7546_{\pm0.0017}}$ & \cellcolor{my_gray} $\mathbf{0.6863_{\pm0.0011}}$ & \cellcolor{my_gray} $0.4640_{\pm0.0005}$ & \cellcolor{my_gray} $\mathbf{0.7604_{\pm0.0005}}$ \\

		& \cellcolor{my_gray}  FedCD(FedAvg) & \cellcolor{my_gray} $0.6786_{\pm0.0003}$ & \cellcolor{my_gray} $\mathbf{0.4652_{\pm0.0007}}$ & \cellcolor{my_gray} $\underline{0.7566_{\pm0.0050}}$ & \cellcolor{my_gray} $\underline{0.6800_{\pm0.0016}}$ & \cellcolor{my_gray} $\underline{0.4628_{\pm0.0006}}$ & \cellcolor{my_gray} $\underline{0.7539_{\pm0.0013}}$ & \cellcolor{my_gray} $\underline{0.6819_{\pm0.0018}}$ & \cellcolor{my_gray} $\underline{0.4636_{\pm0.0001}}$ & \cellcolor{my_gray} $\underline{0.7596_{\pm0.0026}}$ \\
        
		& \cellcolor{my_gray}  FedCD(FedAtt) & \cellcolor{my_gray} $\underline{0.6796_{\pm0.0017}}$ & \cellcolor{my_gray} $\underline{0.4652_{\pm0.0007}}$ & \cellcolor{my_gray} $0.7532_{\pm0.0049}$ & \cellcolor{my_gray} $0.6781_{\pm0.0022}$ & \cellcolor{my_gray} $0.4636_{\pm0.0010}$ & \cellcolor{my_gray} $0.7508_{\pm0.0012}$ & \cellcolor{my_gray} $0.6818_{\pm0.0017}$ & \cellcolor{my_gray} $\mathbf{0.4636_{\pm0.0003}}$ & \cellcolor{my_gray} $0.7562_{\pm0.0058}$ \\

		\hline
  
		\multirow{9}[0]{*}{\textbf{NCD}} 
        & Base   & $0.7227_{\pm0.0008}$ & $0.4512_{\pm0.0077}$ & $\mathbf{0.7880_{\pm0.0013}}$ & $0.7296_{\pm0.0016}$ & $0.4464_{\pm0.0026}$ & $\mathbf{0.7972_{\pm0.0017}}$ & $0.7326_{\pm0.0024}$ & $0.4474_{\pm0.0018}$ & $\mathbf{0.7989_{\pm0.0009}}$ \\
		& FedAvg & $0.6916_{\pm0.0015}$ & $0.4535_{\pm0.0006}$ & $0.7591_{\pm0.0013}$ & $0.6990_{\pm0.0013}$ & $0.4524_{\pm0.0005}$ & $0.7650_{\pm0.0002}$ & $0.7010_{\pm0.0024}$ & $0.4515_{\pm0.0008}$ & $0.7688_{\pm0.0017}$ \\
		& FedAtt & $0.6891_{\pm0.0010}$ & $0.4580_{\pm0.0004}$ & $0.7532_{\pm0.0004}$ & $0.6965_{\pm0.0017}$ & $0.4566_{\pm0.0006}$ & $0.7591_{\pm0.0008}$ & $0.7000_{\pm0.0009}$ & $0.4543_{\pm0.0006}$ & $0.7641_{\pm0.0005}$ \\

		& LoAdaBoost & $0.6973_{\pm0.0006}$ & $0.4642_{\pm0.0005}$ & $0.7693_{\pm0.0006}$ & $0.7062_{\pm0.0011}$ & $0.4546_{\pm0.0009}$ & $0.7761_{\pm0.0006}$ & $0.7050_{\pm0.0011}$ & $0.4546_{\pm0.0006}$ & $0.7771_{\pm0.0004}$ \\
        
            & F2MF & $0.6925_{\pm0.0007}$ & $0.4522_{\pm0.0005}$ & $0.7603_{\pm0.0009}$ & $0.7002_{\pm0.0015}$ & $0.4515_{\pm0.0004}$ & $0.7664_{\pm0.0004}$ & $0.7015_{\pm0.0007}$ & $0.4498_{\pm0.0008}$ & $0.7706_{\pm0.0005}$ \\

            & PFedRec &$0.6778_{\pm0.0003}$ & $0.4686_{\pm0.0011}$ & $0.7659_{\pm0.0001}$ & $0.6734_{\pm0.0013}$ & $0.4693_{\pm0.0011}$ & $0.7764_{\pm0.0006}$ & $0.6714_{\pm0.0023}$ & $0.4693_{\pm0.0051}$ & $0.7746_{\pm0.0037}$ \\
        
            & \cellcolor{my_gray}  FedCD & \cellcolor{my_gray} $\mathbf{0.7283_{\pm0.0009}}$ & \cellcolor{my_gray} $\mathbf{0.4442_{\pm0.0023}}$ & \cellcolor{my_gray} $\underline{0.7788_{\pm0.0011}}$ & \cellcolor{my_gray} $\mathbf{0.7347_{\pm0.0006}}$ & \cellcolor{my_gray} $\underline{0.4402_{\pm0.0029}}$ & \cellcolor{my_gray} $\underline{0.7909_{\pm0.0021}}$ & \cellcolor{my_gray} $\mathbf{0.7374_{\pm0.0010}}$ & \cellcolor{my_gray} $\mathbf{0.4335_{\pm0.0013}}$ & \cellcolor{my_gray} $\underline{0.7934_{\pm0.0023}}$ \\

		& \cellcolor{my_gray}  FedCD(FedAvg) & \cellcolor{my_gray} $\underline{0.7236_{\pm0.0007}}$ & \cellcolor{my_gray} $\underline{0.4490_{\pm0.0011}}$ & \cellcolor{my_gray} $0.7764_{\pm0.0013}$ & \cellcolor{my_gray} $0.7335_{\pm0.0014}$ & \cellcolor{my_gray} $\mathbf{0.4395_{\pm0.0012}}$ & \cellcolor{my_gray} $0.7898_{\pm0.0015}$ & \cellcolor{my_gray} $\underline{0.7349_{\pm0.0005}}$ & \cellcolor{my_gray} $0.4385_{\pm0.0019}$ & \cellcolor{my_gray} $0.7913_{\pm0.0014}$ \\

		& \cellcolor{my_gray}  FedCD(FedAtt) & \cellcolor{my_gray} $0.7227_{\pm0.0007}$ & \cellcolor{my_gray} $0.4490_{\pm0.0015}$ & \cellcolor{my_gray} $0.7750_{\pm0.0014}$ & \cellcolor{my_gray} $\underline{0.7342_{\pm0.0007}}$ & \cellcolor{my_gray} $0.4404_{\pm0.0017}$ & \cellcolor{my_gray} $0.7890_{\pm0.0002}$ & \cellcolor{my_gray} $0.7339_{\pm0.0007}$ & \cellcolor{my_gray} $\underline{0.4383_{\pm0.0019}}$ & \cellcolor{my_gray} $0.7927_{\pm0.0008}$ \\

		\hline
  
		\multirow{9}[0]{*}{\textbf{KaNCD}} 
        & Base & $\mathbf{0.7374_{\pm0.0018}}$ & $\mathbf{0.4281_{\pm0.0049}}$ & $\mathbf{0.8081_{\pm0.0011}}$ & $\mathbf{0.7431_{\pm0.0009}}$ & $\mathbf{0.4247_{\pm0.0015}}$ & $\mathbf{0.8139_{\pm0.0013}}$ & $\mathbf{0.7419_{\pm0.0020}}$ & $\mathbf{0.4264_{\pm0.0026}}$ & $\mathbf{0.8134_{\pm0.0011}}$ \\
        
		& FedAvg & $0.7071_{\pm0.0013}$ & $0.5369_{\pm0.0014}$ & $0.7634_{\pm0.0012}$ & $0.7085_{\pm0.0031}$ & $0.5350_{\pm0.0018}$ & $0.7641_{\pm0.0028}$ & $0.7045_{\pm0.0018}$ & $0.5391_{\pm0.0015}$ & $0.7660_{\pm0.0031}$ \\
        
		& FedAtt & $0.7082_{\pm0.0014}$ & $0.5356_{\pm0.0011}$ & $0.7619_{\pm0.0021}$ & $0.7120_{\pm0.0022}$ & $0.5327_{\pm0.0026}$ & $0.7630_{\pm0.0015}$ & $0.7093_{\pm0.0027}$ & $0.5349_{\pm0.0019}$ & $0.7636_{\pm0.0020}$ \\
        
		& LoAdaBoost & $0.7179_{\pm0.0017}$ & $0.5260_{\pm0.0010}$ & $\underline{0.7749_{\pm0.0008}}$ & $0.7171_{\pm0.0014}$ & $0.5261_{\pm0.0013}$ & $\underline{0.7787_{\pm0.0015}}$ & $0.7131_{\pm0.0011}$ & $0.5299_{\pm0.0011}$ & $\underline{0.7774_{\pm0.0021}}$ \\
            & F2MF & $0.7109_{\pm0.0012}$ & $0.5341_{\pm0.0006}$ & $0.7649_{\pm0.0015}$ & $0.7098_{\pm0.0019}$ & $0.5349_{\pm0.0021}$ & $0.7676_{\pm0.0020}$ & $0.7085_{\pm0.0034}$ & $0.5351_{\pm0.0024}$ & $0.7673_{\pm0.0041}$ \\
            & PFedRec &$0.6939_{\pm0.0019}$ & $0.5461_{\pm0.0015}$ & $0.7614_{\pm0.0008}$ & $0.6821_{\pm0.0120}$ & $0.5575_{\pm0.0115}$ & $0.7695_{\pm0.0015}$ & $0.6833_{\pm0.0082}$ & $0.5581_{\pm0.0076}$ & $0.7711_{\pm0.0006}$ \\

            & \cellcolor{my_gray}  FedCD & \cellcolor{my_gray} $\underline{0.7255_{\pm0.0028}}$ & \cellcolor{my_gray} $\underline{0.5211_{\pm0.0029}}$ & \cellcolor{my_gray} $0.7556_{\pm0.0043}$ & \cellcolor{my_gray} $\underline{0.7256_{\pm0.0019}}$ & \cellcolor{my_gray} $0.5211_{\pm0.0018}$ & \cellcolor{my_gray} $0.7536_{\pm0.0019}$ & \cellcolor{my_gray} $\underline{0.7244_{\pm0.0022}}$ & \cellcolor{my_gray} $0.5222_{\pm0.0021}$ & \cellcolor{my_gray} $0.7527_{\pm0.0030}$ \\
            
		& \cellcolor{my_gray}  FedCD(FedAvg) & \cellcolor{my_gray} $0.7226_{\pm0.0014}$ & \cellcolor{my_gray} $0.5242_{\pm0.0019}$ & \cellcolor{my_gray} $0.7677_{\pm0.0031}$ & \cellcolor{my_gray} $0.7245_{\pm0.0026}$ & \cellcolor{my_gray} $0.5215_{\pm0.0021}$ & \cellcolor{my_gray} $0.7693_{\pm0.0042}$ & \cellcolor{my_gray} $0.7227_{\pm0.0016}$ & \cellcolor{my_gray} $0.5228_{\pm0.0020}$ & \cellcolor{my_gray} $0.7689_{\pm0.0068}$ \\

		& \cellcolor{my_gray}  FedCD(FedAtt) & \cellcolor{my_gray} $0.7213_{\pm0.0020}$ & \cellcolor{my_gray} $0.5214_{\pm0.0044}$ & \cellcolor{my_gray} $0.7683_{\pm0.0091}$ & \cellcolor{my_gray} $0.7254_{\pm0.0011}$ & \cellcolor{my_gray} $\underline{0.5203_{\pm0.0018}}$ & \cellcolor{my_gray} $0.7648_{\pm0.0067}$ & \cellcolor{my_gray} $0.7240_{\pm0.0013}$ & \cellcolor{my_gray} $\underline{0.5219_{\pm0.0016}}$ & \cellcolor{my_gray} $0.7676_{\pm0.0059}$ \\

        \hline

        \multirow{1}[0]{*}{\textbf{GUM}} 
        & HPFL & $0.6864_{\pm0.0021}$ & $0.5195_{\pm0.0020}$ & $0.7401_{\pm0.0027}$ & $0.6978_{\pm0.0016}$ & $0.5142_{\pm0.0016}$ & $0.7502_{\pm0.0011}$ & $0.7026_{\pm0.0005}$ & $0.5123_{\pm0.0006}$ & $0.7505_{\pm0.0015}$ \\
        
        \hline

		\toprule
		\toprule
	\end{tabular}%
    
    \begin{tabular}{l|lr|lr|lr|lr|lr|lr}
    \toprule
    \textbf{Number of times FedCD wins X} & vs. base \quad &  15 \quad & vs. FedAvg \quad & 24 \quad & vs. FedAtt \quad & 24 \quad & vs. LoAdaBoost \quad & 24 \quad & vs. F2MF \quad & 24 \quad & vs. PFedRec \quad & 24 \quad \\

    \bottomrule
    \end{tabular}%

	\label{tab7}%
\end{table*}%

To answer \textbf{RQ1}, FedCD was compared with \textit{Base}, FedAvg, FedAtt, LoAdaBoost, F2MF, and PFedRec under DINA, NCD, and KaNCD on three datasets: ASSIST2009, ASSIST2012, and SLP-Math, where \textit{Base} denotes the centralized training paradigm. Tables~\ref{tab2} and~\ref{tab3} present their performance on ASSIST2009 and ASSIST2012 regarding AUC, ACC, and RMSE obtained under three splitting settings. 
Only the best result among the comparisons is shown in bold, and the second-best result is underlined. As a result, there are $3 \times 3 \times 3 \times 2 = 54$ comparison results between FedCD and each baseline in both tables.

As we can observe, it is intuitive and reasonable that the proposed FedCD framework performs slightly worse than the centralized training base models; however, it exhibits better overall performance than the compared FL approaches in most cases. Specifically, the last rows of both tables indicate how many times FedCD outperforms the other approaches. It can be seen that FedCD beats FedAvg, FedAtt, LoAdaBoost, F2MF, and PFedRec 50, 52, 48, 50, and 37 times out of 54 comparisons, respectively. The NCD-based FedCD surpasses HPFL in each metric, indirectly validating its effectiveness in addressing the client fairness issue. Additionally, compared to the two variants, the overall performance of FedCD is slightly better than \textit{FedCD(Avg)} and \textit{FedCD(Att)}, confirming the effectiveness of the utilized fairness-aware aggregation mechanism.

To further evaluate the effectiveness of FedCD compared to other FL approaches under different CDMs, 
the overall comparison results on the SLP-Math dataset are summarized in 
Table \ref{tab7}, which used the same settings as the experiments on ASSIST2009 and ASSIST2012.

As shown in Table \ref{tab7}, FedCD and its two variants, FedCD(FedAvg) and FedCD(FedAtt), outperformed other FL approaches. 
Specifically, FedCD outperformed FedAvg, FedAtt, LoAdaBoost, F2MF, and PFedRec by 24, 24, 24, 24, and 24 times out of a total of 27 comparisons, respectively. 
Furthermore, the performance of FedCD based on NCD and DINA even surpasses that of the centralized model, demonstrating that the proposed aggregation strategy can enhance the overall performance of the base models.

\subsection{Effectiveness of Parameter Decoupling based Personalization Strategy (RQ2)}

\subsubsection{Parameter Decoupling Effectiveness}

The comparisons between FedCD and \textit{FedCD(Avg)} as well as \textit{FedCD(Att)} in Tables~\ref{tab2} and~\ref{tab3} have demonstrated the effectiveness of the fairness-aware aggregation mechanism in the proposed parameter decoupling-based personalization strategy to some extent.

\begin{table}[t]
  \centering
  \caption{Performance of FedCD and its six variants on two datasets based on  NCD under the ratio of 80\%/20\%}
      \renewcommand{\arraystretch}{1.1}
	\setlength{\tabcolsep}{1.mm}
 
    \begin{tabular}{l|rrr|rrr}
    \toprule
    \multicolumn{1}{c|}{\multirow{2}[4]{*}{\textbf{Models}}} & \multicolumn{3}{c|}{\textbf{ASSIST2009}} & \multicolumn{3}{c}{\textbf{ASSIST2012}} \\
\cmidrule{2-7}          & \multicolumn{1}{l}{\textbf{ACC}} & \multicolumn{1}{l}{\textbf{RMSE}} & \multicolumn{1}{l|}{\textbf{AUC}} & \multicolumn{1}{l}{\textbf{ACC}} & \multicolumn{1}{l}{\textbf{RMSE}} & \multicolumn{1}{l}{\textbf{AUC}} \\
    \midrule
    {FedAvg} & 67.99  & 46.01  & 67.20  & 71.13  & 44.71  & 67.23  \\
    \textit{FedCD(Avg)-p.d.p.} & 68.49  & 46.50  & 71.87  & 71.56  & 44.99  & 70.83  \\
    \textit{FedCD(Avg)} & 70.29  & 44.63  & 73.15  & 71.76  & 44.08  & 71.44  \\
      \hline
    {FedAtt} & 67.61  & 46.94  & 66.55  & 71.13  & 44.89  & 67.16  \\
    \textit{FedCD(Att)-p.d.p.} & 68.01  & 46.89  & 71.33  & 71.46  & 44.98  & 70.85  \\
    \textit{FedCD(Att)} & 70.30  & 44.69  & 73.62  & 71.58  & 44.39  & 71.48  \\
    \hline
    \textbf{FedCD} & 71.43  & 44.09  & 74.15  & 71.94  & 44.70  & 70.43  \\
    \bottomrule
    \end{tabular}%
  \label{tab4}%
\end{table}%

To further explore the effectiveness of this strategy, 
we created two variants of the proposed FedCD, denoted as \textit{FedCD(Avg)-p.d.p.}, and   \textit{FedCD(Att)-p.d.p.}, with detailed descriptions summarized in \textbf{Section 5.1.2.}
Table~\ref{tab4} presents the performance of FedCD and its six variants on two datasets using NCD as the base model. 
As shown in Table~\ref{tab4}, the comparisons between \textit{FedCD(Avg)} and \textit{FedCD(Avg)-p.d.p.} confirm the effectiveness of the role of decoupling parameters for personalization in the strategy.
 The same conclusion can be drawn from the results of the \textit{FedCD(Att)} series.

\begin{figure*}[t]
\centering
\includegraphics[width=0.8\linewidth]{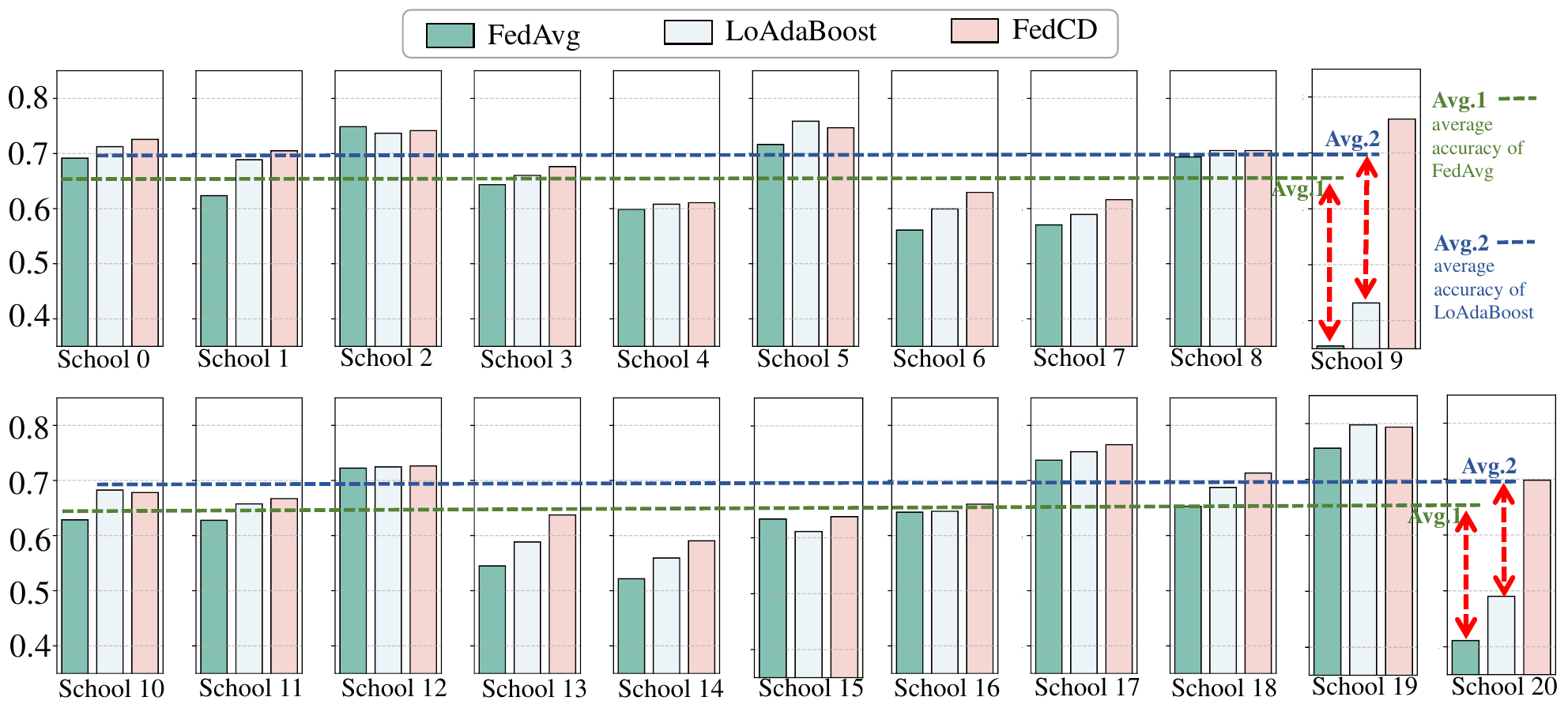} 
\caption{Each client's ACC values of FedCD, FedAvg, and LoAdaBoost on ASSIST2009 under NCD.}
\label{fig3}
\end{figure*}

\begin{table}[t]
  \centering
      \renewcommand{\arraystretch}{1.1}
	\setlength{\tabcolsep}{1.mm}
  \caption{Each client's statistics and predicted ACC}
    \begin{tabular}{l|rrcrcrr}
    \toprule
    \textbf{Statistics/} & \multicolumn{1}{l}{\textbf{School 0}} & \multicolumn{1}{l}{\textbf{ 1}} & \textbf{…} &  \cellcolor{my_gray}  \textbf{9} & \textbf{…} & \textbf{19} &  \cellcolor{my_gray} 
 \textbf{20} \\
    \midrule
    \textbf{\# of students} & 205 & 306 & \multirow{5}[2]{*}{…} &  \cellcolor{my_gray}  150 & \multirow{5}[2]{*}{…}    & 74 &  \cellcolor{my_gray}  65\\
    \textbf{\# of wrongs} & 7176  & 7767  &       &  \cellcolor{my_gray}  4003  &         & 259 &  \cellcolor{my_gray}  536 \\
    \textbf{\# of corrects} & 15827 & 10408 &       &  \cellcolor{my_gray}  1642  &          & 818 &  \cellcolor{my_gray}  327\\
    \textbf{\# of logs} & 23003 & 18175 &       &  \cellcolor{my_gray}  5645  &         & 1077 &  \cellcolor{my_gray}  863 \\
    \textbf{Correct rate} & 0.688  & 0.573  &       & \cellcolor{my_gray}  0.291  &        & 0.759  &  \cellcolor{my_gray}  0.379 \\
    \midrule
    \textbf{FedAvg} & 69.15  & 62.35  & \multirow{3}[2]{*}{…} &  \cellcolor{my_gray}  35.46  & \multirow{3}[2]{*}{…}   & 75.60  & \cellcolor{my_gray}  41.09\\
    \textbf{LoAdaBoost} & 71.24  & 68.87  &       &  \cellcolor{my_gray}  43.17  &         & 79.76  &  \cellcolor{my_gray}  49.01\\
    \textbf{FedCD} & 72.55  & 70.47  &       &  \cellcolor{my_gray}  76.05  &         & 79.37  &  \cellcolor{my_gray}  69.80\\
    \bottomrule
    \end{tabular}%
  \label{tab5}

\end{table}%

\subsubsection{The Influence of Private Exercise Embeddings}

In model design of FedCD, the diagnostic function is personalized for each client, while the exercise embeddings are globally shared. To isolate the effect of the global exercise embeddings on fairness, we modified the model to make the exercise embeddings also client-specific, and denoted this variant as FedCD\_per. We randomly selected a subset of schools whose performance was below the average ACC values of schools, and conducted the experiment with NCD-based models on ASSIST2009. 

\begin{table}[htbp]
  \centering
        \renewcommand{\arraystretch}{1.1}
  \caption{Performance of schools with applying private exercise embeddings into FedCD on ASSIST2009}
  \label{tab_priexe}%
    \begin{tabular}{l|rrrrr|r}
    \toprule
    \textbf{Model\textbackslash{}Schoool }& \textbf{4    } & \textbf{7   }  & \textbf{8  }   & \textbf{10 }   &\textbf{ 14 }   & \multicolumn{1}{l}{\textbf{Average}} \\
    \midrule
    \textbf{FedCD\_per} & 0.5883 & 0.6051 & 0.6974 & 0.672 & 0.596 & 0.7036 \\
    \midrule
    \textbf{FedCD} & \textbf{0.6338} & \textbf{0.6351} & \textbf{0.7367} & \textbf{0.7007} & \textbf{0.6582} & \textbf{0.7092} \\
    \bottomrule
    \end{tabular}%
\end{table}%

As shown in Table~\ref{tab_priexe}, receiving global information through shared exercise embeddings contributes to fairness and improves overall performance. However, its contribution to fairness is not as substantial as the personalization of the diagnostic function, but it still brings effective improvements to alleviate fairness issues.

\subsection{Effectiveness in Client Fairness (RQ3)}

\begin{figure*}[t]
\centering
\includegraphics[width=1\linewidth]{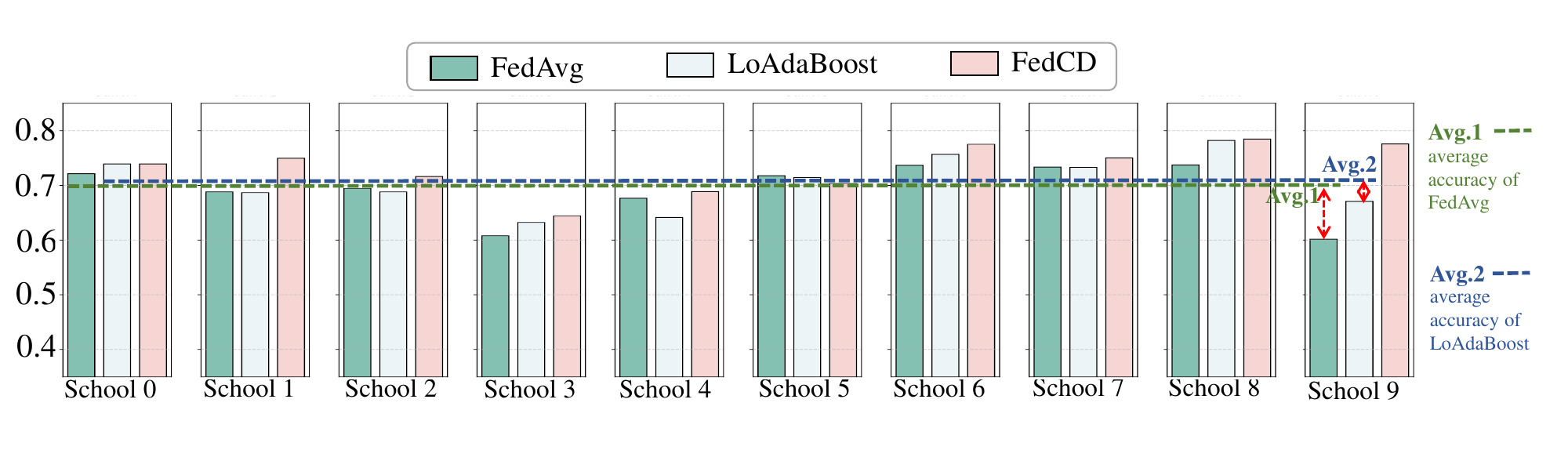} 
\caption{Each client's ACC values of FedCD, FedAvg, and LoAdaBoost on the SLP-Math dataset under the NCD base model.}
\label{fig4}
\end{figure*}

To answer \textbf{RQ3}, Fig.~\ref{fig3} shows the ACC values for each client/school obtained by FedAvg, LoAdaBoost, and FedCD under the NCD base model on ASSIST2009 and SLP-Math. It is evident that FedCD maintains a certain level of accuracy advantage over FedAvg and LoAdaBoost for each client/school, with this advantage being particularly significant in Schools 9 and 20. For a deeper insight into the results, Table~\ref{tab5} summarizes the statistics of each client's student learning data in the upper part and presents each client's ACC for FedAvg, LoAdaBoost, and FedCD on ASSIST2009 in the lower part. 
It can be inferred that the significantly lower ACC values of FedAvg and LoAdaBoost in Schools 9 and 20 are not influenced by the number of students or the number of logs contained in each client. The unusually low ACC values for these two clients compared to others may be attributed to their lower correct response rates.

Therefore, these two clients do not face a data imbalance issue but rather a client fairness issue, where the FL models perform unfairly on clients with abnormal data quality that is out of distribution. Nevertheless, the proposed FedCD achieves similar ACC values for Schools 9 and 20 as for other schools, indicating that FedCD effectively addresses the client fairness issue.

To further investigate the effectiveness of FedCD in addressing the client fairness issue, we conducted additional experiments using FedAvg, LoAdaBoost, and FedCD based on NCD on the SLP-Math dataset. 
Fig~\ref{fig4} illustrates the ACC values of the three approaches across different schools.

The results, which are largely consistent with the conclusions drawn in \textbf{RQ3}, clearly demonstrate that FedCD achieves varying degrees of improvement in ACC across all clients. 
Notably, the enhancement is most pronounced for School 9, where the ACC is significantly higher than the average values attained by FedAvg and LoAdaBoost. This improvement can be attributed to the fact that School 9 has a larger proportion of negative samples, a challenge that FedCD effectively mitigates.

The following Table~\ref{tab_ass12} shows additional experiments on ASSIST2012. We randomly selected 10 schools for the experimental results, which are consistent with the performance on the other two datasets, proving that FedCD can alleviate the fairness issue.

\begin{table}[htbp]
  \centering
        \renewcommand{\arraystretch}{1.1}
  \caption{Each client’s ACC values of FedCD, FedAvg, and LoAdaBoost on the ASSIST2012 dataset under the KaNCD base model}
  \label{tab_ass12}
    \begin{tabular}{l|rrrrrrrrrr}
    \toprule
    \textbf{Model}\textbackslash{}\textbf{Schoool} & \textbf{0 }    & \textbf{1}     & \textbf{2}     & \textbf{3}     & \textbf{4}     & \textbf{5}     & \textbf{6}     & \textbf{7}     & \textbf{8}     & \textbf{9} \\
    \midrule
    \textbf{FedAvg} & 0.5810 & 0.6551 & 0.6143 & 0.6799 & 0.6343 & 0.6490 & 0.7002 & 0.5504 & 0.6422 & 0.7111 \\
    \textbf{LoAdaBoost} & 0.6913 & 0.6543 & 0.6742 & 0.6999 & 0.6633 & 0.6357 & 0.6829 & 0.4822 & 0.6238 & 0.6659 \\
    \midrule
    \textbf{FedCD} & \textbf{0.7557} & \textbf{0.6983} & \textbf{0.7213} & \textbf{0.7122} & \textbf{0.7204} & \textbf{0.6642} & \textbf{0.7273} & \textbf{0.6217} & \textbf{0.6654} & \textbf{0.7494} \\
    \bottomrule
    \end{tabular}%
\end{table}%

In conclusion, the effectiveness of the proposed FedCD in addressing the client fairness issue is strongly validated, highlighting its potential as a robust solution in FL scenarios.

\subsection{Sensitivity of Key Hyperparameters (RQ4)}

\subsubsection{Sensitivity Analysis of Coefficient}

As highlighted, an important hyperparameter $\gamma$ in Eq.\ref{eq_exer} represents client importance. 
To investigate its impact on the performance of the proposed FedCD, we conducted a sensitivity analysis on the ASSIST2009 and SLP-Math datasets. Specifically, we fixed all other parameters in FedCD and varied $\gamma$ across a range of values from 0.1 to 0.6, with increments of 0.1.

\begin{figure*}[htbp]
\centering
\includegraphics[width=1.\linewidth]{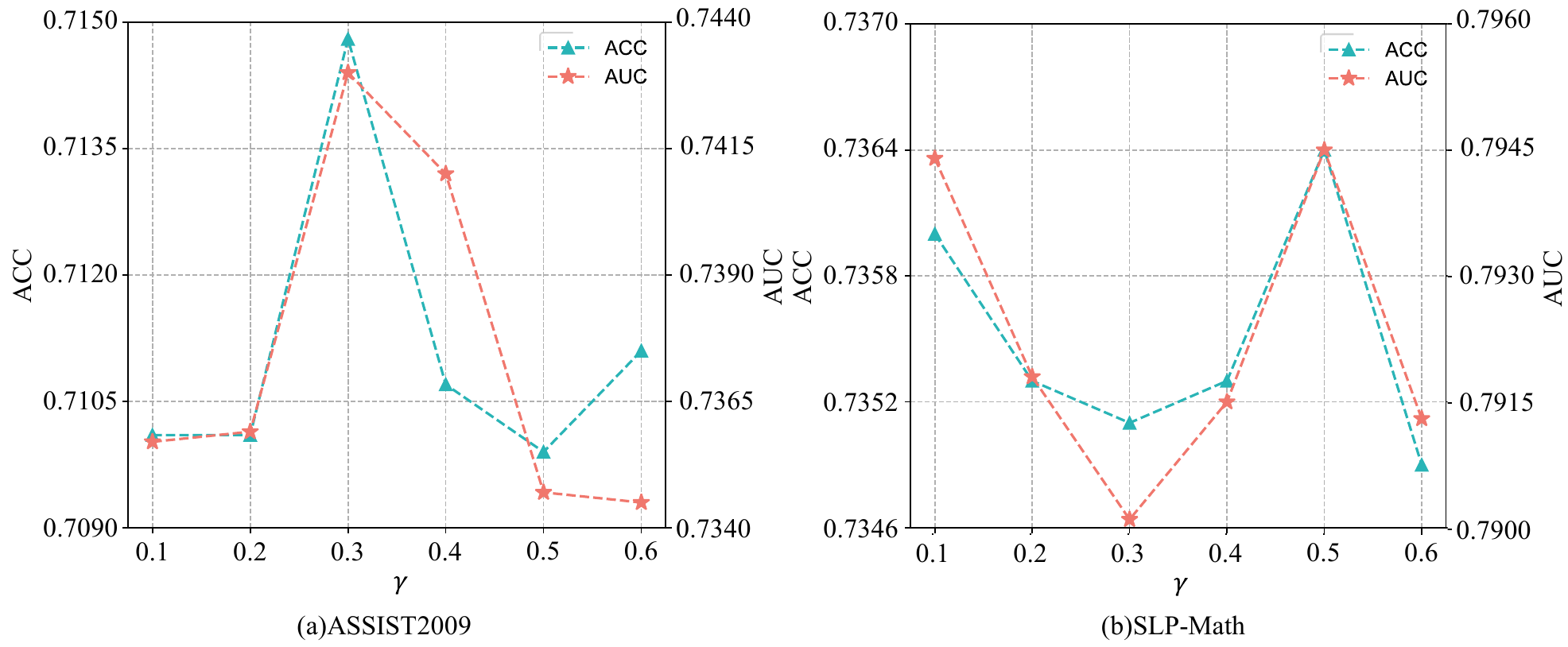} 
\caption{Sensitivity analysis of FedCD to coefficient $\gamma$ on the ASSIST2009 and SLP-Math datasets.}
\label{fig5}
\end{figure*}

As illustrated in Fig.~\ref{fig5}, the proposed FedCD demonstrated different performance on the two datasets due to the differing data distributions among clients in ASSIST2009 and SLP-Math. Notably, the best results were achieved when $\gamma$ was set to 0.3 for ASSIST2009 and 0.5 for SLP-Math, respectively. 
However, when the coefficient became too large, the model tended to favor better-performing clients while neglecting those with poorer performance, leading to a decline in overall effectiveness. 
To strike a balance across the three datasets, we adopted $\gamma = 0.1$, which ensures promising performance across all datasets.

\subsubsection{The Influence of Personalization Ratio of Diagnostic Function Parameters }

In FedCD, all parameters of the diagnostic function are personalized. To investigate the impact of varying degrees of parameter personalization on model performance, we defined a personalized-to-global parameter ratio and conducted experiments using FedCD based on the NCD model on the ASSIST2009 dataset.
With NCD, three variants of FedCD were created: FedCD(0\%) with all parameters globally shared, FedCD(w/o W\_fc) with parameters of three-layer FC globally shared, and FedCD(w/o W\_disc) with parameters of W\_disc globally shared.

\begin{table}[htbp]
  \centering
        \renewcommand{\arraystretch}{1.1}
  \caption{Different personalization ratio of FedCD on the ASSIST2009 dataset under the NCD base model}
  \label{tab_perratio}
    \begin{tabular}{l|r|r|r}
    \toprule
    & \multicolumn{1}{l|}{\textbf{ACC}} & \multicolumn{1}{l|}{\textbf{RMSE}} & \multicolumn{1}{l}{\textbf{AUC}} \\
    \midrule
    \textbf{FedCD(0\%)} & 0.7114 & 0.4587 & 0.7221 \\
   \textbf{ FedCD(w/o W\_fc) }& 0.7151 & 0.456 & 0.7383 \\
    \textbf{FedCD(w/o W\_disc)} & 0.7184 & 0.4453 & 0.7389 \\
    \midrule
    \textbf{FedCD} & \textbf{0.7201} & \textbf{0.4439} & \textbf{0.7413} \\
    \bottomrule
    \end{tabular}%
\end{table}%

Obviously, FedCD holds better performance with more personalized parameters, which means that more parameters can better learn the personalized features of local data.

The results in Table~\ref{tab_perratio} reveal that increasing the degree of personalization consistently improved model performance across all metrics. Full personalization in FedCD yielded the best results, achieving the best performance. This demonstrated that selectively personalizing interaction parameters significantly enhanced the model's ability to adapt to heterogeneous client data, validating the effectiveness of personalized parameter design in FedCD.

\subsubsection{Sensitivity Analysis of Local Epoch}

In federated learning, the number of local epochs plays a critical role in determining client performance. A smaller number of local epochs often results in undertraining, thereby degrading overall performance, while a larger number increased training time and posed a risk of overfitting. To identify the optimal setting, we conducted experiments on the ASSIST2009 dataset by varying the number of local epochs in FedCD from 1 to 5.

\begin{table}[htbp]
  \centering
        \renewcommand{\arraystretch}{1.1}
  \caption{Performance of FedCD with various local epoch on ASSIST2009}
  \label{tab_epo}%
    \begin{tabular}{l|rrrrr}
    \toprule
    \textbf{Local\_epoch} &\textbf{ 1}     & \textbf{2  }   & \textbf{3 }    &\textbf{ 4 }    &\textbf{ 5} \\
    \midrule
    \textbf{FedCD} & 0.7017 & 0.7142 & 0.7077 & 0.7039 & 0.7201 \\
    \bottomrule
    \end{tabular}%
\end{table}%

The results in Table~\ref{tab_epo} show that FedCD maintains stable performance across different local training epochs, with the best ACC achieved at local\_epoch 5. This indicates that moderate increases in local updates can enhance model performance without overfitting.

\subsection{Effectiveness regarding Privacy Protection (RQ5)}

To further enhance the privacy protection capability of FedCD using the local differential privacy strategy, we evaluated it on ASSIST2009 under the NCD with the intensity $\delta$ set to 0, 0.10, 0.20, 0.30, 0.40, and 0.50. Table \ref{tab6} summarizes the results.

\begin{table}[htbp]
  \caption{Performance of applying local differential privacy technique into FedCD with various    $\delta$ on ASSIST2009}
  \label{tab6}
      \renewcommand{\arraystretch}{1.1}
  \begin{tabular}{llllll}
    \toprule
   \textbf{ Intensity $\delta$ }
          & \textbf{0.10}  & \textbf{0.20}  & \textbf{0.30} & \textbf{0.40} & \textbf{0.50}\\
    \midrule
    \textbf{ACC}   & 70.89 & 70.96 & 70.13 & 69.53 & 68.11\\
    \textbf{RMSE}  & 46.18 & 46.11 & 46.07 & 46.15 & 46.07\\
    \textbf{AUC }  & 72.96 & 71.22 & 70.76 & 70.62 & 69.47\\
    \bottomrule
  \end{tabular}
\end{table}

The results in Table \ref{tab6} show a slight decline in both ACC and AUC as $\delta$ increases, while RMSE remains relatively stable. In particular, performance degradation becomes more apparent when $\delta$ exceeds 0.30. This suggests that although privacy protection is enhanced with higher noise, it comes at the cost of reduced model accuracy. Hence, selecting a moderate $\delta$ is crucial to maintain an effective balance between privacy and utility.


\subsection{Further Discussion of FedCD (RQ6)} 

\subsubsection{Fairness Verification}

The experiment in \textbf{RQ2} has effectively proved that FedCD can improve the performance of clients with poor performance, to further validate whether FedCD could address performance disparity among clients and improve overall fairness.
We adopted group fairness(GF)~\cite{zhang2024towards}, a widely used metric in recommendation systems, as our fairness criterion.

\begin{equation}
\label{eq_fair}
G F=\left|\frac{1}{m_{0}} \sum_{s_{i} \in G_{0}} \mathcal{M}\left(s_{i}\right)-\frac{1}{m_{1}} \sum_{s_{i} \in G_{1}} \mathcal{M}\left(s_{i}\right)\right|,
\end{equation}				
Let $\mathcal{M}$ denote the performance metric, where we used ACC. Clients were divided into two groups based on the mean ACC, and the GF value was calculated accordingly. To achieve fairness among schools, our goal is to minimize GF. The experiment was conducted on the ASSIST2009 dataset, comparing FedCD with two baseline methods: FedAvg and F2MF, both based on NCD.

\begin{table}[htbp]
  \centering
        \renewcommand{\arraystretch}{1.1}
  \caption{GF and ACC values of FedCD, FedAvg, and F2MF on the ASSIST2009 dataset under the NCD base model}
  \label{tab_fair}
  \begin{tabular}{l|c|c}
    \toprule
    \textbf{Model} & \textbf{GF} & \textbf{ACC} \\
    \midrule
    \textbf{FedAvg} & 0.167 & 0.680 \\
    \textbf{F2MF} & 0.158 & 0.682 \\
    \textbf{FedCD} & \textbf{0.084} & \textbf{0.720} \\
    \bottomrule
  \end{tabular}
\end{table}
		
The results in Table~\ref{tab_fair} demonstrate that FedCD significantly improved both ACC and fairness across clients. Specifically, it achieved the highest ACC while also exhibiting the lowest GF, indicating reduced disparity among client performances. Compared to FedAvg and F2MF, which have a higher GF and a lower ACC, FedCD effectively balances the utility of the model with an equitable performance distribution. 

\subsubsection{Consistency Evaluation}

Intuitively, if student $ a $ a demonstrates a higher level of mastery over knowledge concept $ k $ compared to student $b$, then student $ a $ is expected to have a greater likelihood of correctly answering exercises that involve concept 
$ k $. To evaluate the ranking performance with respect to such relative proficiency, we adopted the Degree of Agreement (DOA) as the assessment metric. For a given knowledge concept $ k $, the DOA score, denoted as $\operatorname{DOA}(k)$, is defined as follows:

\begin{equation}
\label{eq_doa}
\operatorname{DOA}(k)=\frac{1}{Z} \sum_{a=1}^{N} \sum_{b=1}^{N} \delta\left(F_{a k}^{s}, F_{b k}^{s}\right) \sum_{j=1}^{M} I_{j k} \frac{J(j, a, b) \wedge \delta\left(r_{a j}, r_{b j}\right)}{J(j, a, b)},
\end{equation}	
let $ Z=\sum_{a=1}^{N} \sum_{b=1}^{N} \delta\left(F_{a k}^{s}, F_{b k}^{s}\right)$, where $F_{a k}^{s}$  represents the estimated proficiency of student $ a $ on concept $ k $, and $\delta(x, y)=1$  if  x>y  and $ \delta(x, y)=0 $ otherwise.  
The indicator $I_{j k}=1$  if exercise $ j $ is associated with knowledge concept $k $, and  $I_{j k}=0 $ otherwise. Similarly, $J(j, a, b)=1$  if both student $a$ and $b $ have attempted exercise $j$, and $J(j, a, b)=0$ otherwise. The average of $\operatorname{DOA}(k)$ across all knowledge concepts is employed to measure the overall quality of the diagnostic outcomes, i.e., the accuracy of the knowledge proficiency estimates provided by the model. The experiment was conducted on ASSIST2009 with NCD-based FedAvg, FedAtt, FedRAP, and FedCD.

\begin{table}[htbp]
  \centering
        \renewcommand{\arraystretch}{1.1}
  \caption{DOA and ACC values of FedCD, FedAvg, and F2MF on the ASSIST2009 dataset under the NCD base model}
  \label{tab_doa}
    \begin{tabular}{l|r|r}
    \toprule
    & \multicolumn{1}{l|}{\textbf{DOA}} & \multicolumn{1}{l}{\textbf{ACC}} \\
    \midrule
    \textbf{FedAvg} & 0.557 & 0.680 \\
    \textbf{FedAtt} & 0.560  & 0.675 \\
    \textbf{FedRAP} & 0.602 & 0.701 \\
    \midrule
    \textbf{FedCD} & \textbf{0.612} & \textbf{0.720} \\
    \bottomrule
    \end{tabular}%
\end{table}%

As shown in Table~\ref{tab_doa}, FedCD achieves the highest DOA score, demonstrating superior capability in capturing students' relative proficiencies. This is mainly due to its personalized FL design, which tailors local models to individual student distributions. Unlike global-only approaches, FedCD better preserves proficiency rankings by accounting for learner diversity, leading to improved interpretability and diagnostic quality.

\subsubsection{The Influence of Varying the Loss Function Weights}

In FedCD, to promote fairness, we assigned equal loss weights (set to 1) to all clients, encouraging the model to give equal attention to clients with smaller data volumes. This approach contrasts with previous methods such as FedAvg, which prioritizes clients with larger data volumes and often neglects those with fewer samples. To evaluate the effectiveness of this fairness-enhancing strategy, 
we modified the aggregation of client losses in FedCD by weighting each client's loss according to its proportion of the total data volume. We denoted this variant as FedCD\_loss\_avg, which differs from the FedCD (FedAvg) in our paper, where the parameter aggregation weights are based on data volume. To evaluate the impact, we randomly selected one-fourth of the schools and presented the comparison results. Both FedCD and FedCD\_loss\_avg were implemented based on the NCD model and evaluated on the ASSIST2009 dataset.

\begin{table}[htbp]
  \centering
        \renewcommand{\arraystretch}{1.1}
  \caption{Performance Comparison on schools Between FedCD and FedCD\_loss\_avg on the ASSIST2009 dataset}
   \label{tab_vary}%
    \begin{tabular}{r|r|r|r}
    \toprule
    \multicolumn{1}{l|}{\textbf{School id}} & \multicolumn{1}{l|}{\textbf{Log\_num}} & \multicolumn{1}{l|}{\textbf{FedCD\_loss\_avg}} & \multicolumn{1}{l}{\textbf{FedCD}} \\
    \midrule
    \textbf{2 }    & 99306 & 0.7392 & 0.7337 \\
    \textbf{3  }   & 9766  & 0.6666 & 0.6720 \\
    \textbf{5   }  & 11463 & 0.7613 & 0.7545 \\
    \textbf{9    } & 1085  & 0.6377 & 0.6397 \\
    \textbf{12    }& 4173  & 0.7392 & 0.7333 \\
    \bottomrule
    \end{tabular}%
\end{table}%

As shown in Table~\ref{tab_vary}, FedCD\_loss\_avg and FedCD both perform well on Schools 2, 5, and 12, but since FedCD\_loss\_avg inherently favors clients with larger datasets, its performance on these schools is even better. Conversely, on Schools 3 and 9, where both methods already perform poorly, FedCD\_loss\_avg performs even worse due to underweighting clients with fewer samples.

These results suggest that modifying the loss aggregation weights provides an effective means to promote fairness in federated cognitive diagnosis. As shown in Table~\ref{tab_vary}, FedCD\_loss\_avg and FedCD both perform well on Schools 2, 5, and 12, but since FedCD\_loss\_avg inherently favors clients with larger data volumes, its performance on these schools is even better.
Conversely, on Schools 3 and 9, where both methods already perform poorly, FedCD\_loss\_avg performs even worse due to underweighting clients with fewer samples.

\subsubsection{Statistics of Computational Overhead}

For FedCD, its each round (P rounds) of computational overhead includes training each client (T clients) of models for $Num$ epochs and aggregating exercise parameters (with dimensions $M \cdot D$). $|R|$ denotes training data size. The first part is about  $O(T \cdot N u m \cdot|R| \cdot \hat{d})$, where  $O(\hat{d})$  denotes overhead of one forward and one backward pass; the second is about $O(T \cdot(1+M \cdot D))$. Totally, FedCD holds  $O(P \cdot(T \cdot N u m \cdot|R| \cdot \hat{d}+T \cdot(1+M \cdot D))  )$. Compared to other methods, the different overhead lies in the second part. Although FedCD's second part may be worse than others, it has much less impact on the overhead than the first part.

Similarly, the communication overhead of FedCD is $O(2 \cdot M \cdot D)$, which is lower than that of non-personalized methods such as FedAvg, as it does not require transmitting the entire set of model parameters.

To evaluate the practical time overhead of different FL methods, we selected two representative approaches, FedAvg and FedCD, and conducted experiments on the ASSIST2009 dataset.

\begin{table}[htbp]
  \centering
        \renewcommand{\arraystretch}{1.1}
  \caption{Time consumption of FedCD and FedAvg on the ASSIST2009 dataset under the NCD base model}
  \label{tab_overhead}
    \begin{tabular}{l|r|r}
    \toprule
    \textbf{Seconds} & \multicolumn{1}{l|}{\textbf{FedAvg}} & \multicolumn{1}{l}{\textbf{FedCD}} \\
    \midrule
    \textbf{Train\_time} & 579.76 & 588.18 \\
    \midrule
    \textbf{Evaluation\_time} & 102.15 & 103.94 \\
    \bottomrule
    \end{tabular}%
  \label{tab:addlabel}%
\end{table}%

As shown in Table~\ref{tab_overhead}, although FedCD employs a more complex model structure than FedAvg, the increase in training and evaluation time is less than 10 seconds in total. Despite the slight increase in computational cost, FedCD achieves significantly better performance, demonstrating its efficiency and practicality in real-world FL scenarios.

\section{Conclusion}

This paper proposes a fairness-aware federated cognitive diagnosis framework, called FedCD. The proposed FedCD follows the basic procedures of common FL approaches but adopts a novel parameter decoupling-based personalization strategy. The main idea of this strategy is to divide model parameters into two parts: a locally personalized part and a globally shared part, where the former allows each client to have personalized parameters to learn local data features. For the latter, the fairness-aware parameter aggregation mechanism is used to update the exercise embedding parameters to address the client fairness issue. Compared to five FL approaches across three CDMs on three datasets, the proposed FedCD demonstrates competitive performance, and its capability to address the issue of client fairness has also been validated.

\begin{acks}
This work was supported in part by the National Natural Science Foundation of China (No.62302010),  in part by the Postdoctoral Fellowship Program (Grade B) of China Postdoctoral Science Foundation (No.GZB20240002), and in part by the Anhui Province Key Laboratory of Intelligent Computing and Applications (No. AFZNJS2024KF01).
\end{acks}

\bibliographystyle{ACM-Reference-Format}
\bibliography{sample-base}


\begin{thebibliography}{67}


\ifx \showCODEN    \undefined \def \showCODEN     #1{\unskip}     \fi
\ifx \showISBNx    \undefined \def \showISBNx     #1{\unskip}     \fi
\ifx \showISBNxiii \undefined \def \showISBNxiii  #1{\unskip}     \fi
\ifx \showISSN     \undefined \def \showISSN      #1{\unskip}     \fi
\ifx \showLCCN     \undefined \def \showLCCN      #1{\unskip}     \fi
\ifx \shownote     \undefined \def \shownote      #1{#1}          \fi
\ifx \showarticletitle \undefined \def \showarticletitle #1{#1}   \fi
\ifx \showURL      \undefined \def \showURL       {\relax}        \fi
\providecommand\bibfield[2]{#2}
\providecommand\bibinfo[2]{#2}
\providecommand\natexlab[1]{#1}
\providecommand\showeprint[2][]{arXiv:#2}

\bibitem[Academy(2024)]%
        {Khan}
\bibfield{author}{\bibinfo{person}{Khan Academy}.}
  \bibinfo{year}{2024}\natexlab{}.
\newblock \bibinfo{title}{Khan Academy}.
\newblock
\urldef\tempurl%
\url{https://www.khanacademy.org/}
\showURL{%
\tempurl}


\bibitem[Adnan et~al\mbox{.}(2022)]%
        {adnan2022federated}
\bibfield{author}{\bibinfo{person}{Mohammed Adnan}, \bibinfo{person}{Shivam
  Kalra}, \bibinfo{person}{Jesse~C Cresswell}, \bibinfo{person}{Graham~W
  Taylor}, {and} \bibinfo{person}{Hamid~R Tizhoosh}.}
  \bibinfo{year}{2022}\natexlab{}.
\newblock \showarticletitle{Federated learning and differential privacy for
  medical image analysis}.
\newblock \bibinfo{journal}{\emph{Scientific reports}} \bibinfo{volume}{12},
  \bibinfo{number}{1} (\bibinfo{year}{2022}), \bibinfo{pages}{1953}.
\newblock


\bibitem[Alom et~al\mbox{.}(2018)]%
        {alom2018history}
\bibfield{author}{\bibinfo{person}{Md~Zahangir Alom}, \bibinfo{person}{Tarek~M
  Taha}, \bibinfo{person}{Christopher Yakopcic}, \bibinfo{person}{Stefan
  Westberg}, \bibinfo{person}{Paheding Sidike}, \bibinfo{person}{Mst~Shamima
  Nasrin}, \bibinfo{person}{Brian~C Van~Esesn}, \bibinfo{person}{Abdul A~S
  Awwal}, {and} \bibinfo{person}{Vijayan~K Asari}.}
  \bibinfo{year}{2018}\natexlab{}.
\newblock \showarticletitle{The history began from alexnet: A comprehensive
  survey on deep learning approaches}.
\newblock \bibinfo{journal}{\emph{arXiv preprint arXiv:1803.01164}}
  (\bibinfo{year}{2018}).
\newblock


\bibitem[Chai et~al\mbox{.}(2020)]%
        {chai2020secure}
\bibfield{author}{\bibinfo{person}{Di Chai}, \bibinfo{person}{Leye Wang},
  \bibinfo{person}{Kai Chen}, {and} \bibinfo{person}{Qiang Yang}.}
  \bibinfo{year}{2020}\natexlab{}.
\newblock \showarticletitle{Secure federated matrix factorization}.
\newblock \bibinfo{journal}{\emph{IEEE Intelligent Systems}}
  \bibinfo{volume}{36}, \bibinfo{number}{5} (\bibinfo{year}{2020}),
  \bibinfo{pages}{11--20}.
\newblock


\bibitem[Coursera(2024)]%
        {Coursera}
\bibfield{author}{\bibinfo{person}{Coursera}.} \bibinfo{year}{2024}\natexlab{}.
\newblock \bibinfo{title}{Coursera}.
\newblock
\urldef\tempurl%
\url{https://www.coursera.org/}
\showURL{%
\tempurl}


\bibitem[Cui et~al\mbox{.}(2025)]%
        {cui2025rebalancing}
\bibfield{author}{\bibinfo{person}{Jiajun Cui}, \bibinfo{person}{Hong Qian},
  \bibinfo{person}{Chanjin Zheng}, \bibinfo{person}{Lu Wang},
  \bibinfo{person}{Mo Yu}, {and} \bibinfo{person}{Wei Zhang}.}
  \bibinfo{year}{2025}\natexlab{}.
\newblock \showarticletitle{Rebalancing Discriminative Responses for Knowledge
  Tracing}.
\newblock \bibinfo{journal}{\emph{ACM Transactions on Information Systems}}
  (\bibinfo{year}{2025}).
\newblock


\bibitem[Embretson and Reise(2013)]%
        {embretson2013item}
\bibfield{author}{\bibinfo{person}{Susan~E Embretson} {and}
  \bibinfo{person}{Steven~P Reise}.} \bibinfo{year}{2013}\natexlab{}.
\newblock \bibinfo{booktitle}{\emph{Item response theory}}.
\newblock \bibinfo{publisher}{Psychology Press}.
\newblock


\bibitem[Feng et~al\mbox{.}(2009)]%
        {Assistments09}
\bibfield{author}{\bibinfo{person}{Mingyu Feng}, \bibinfo{person}{Neil
  Heffernan}, {and} \bibinfo{person}{Kenneth Koedinger}.}
  \bibinfo{year}{2009}\natexlab{}.
\newblock \showarticletitle{Addressing the assessment challenge with an online
  system that tutors as it assesses}.
\newblock \bibinfo{journal}{\emph{User modeling and user-adapted interaction}}
  \bibinfo{volume}{19} (\bibinfo{year}{2009}), \bibinfo{pages}{243--266}.
\newblock


\bibitem[Gao et~al\mbox{.}(2021a)]%
        {gao2021rcd}
\bibfield{author}{\bibinfo{person}{Weibo Gao}, \bibinfo{person}{Qi Liu},
  \bibinfo{person}{Zhenya Huang}, \bibinfo{person}{Yu Yin},
  \bibinfo{person}{Haoyang Bi}, \bibinfo{person}{Mu-Chun Wang},
  \bibinfo{person}{Jianhui Ma}, \bibinfo{person}{Shijin Wang}, {and}
  \bibinfo{person}{Yu Su}.} \bibinfo{year}{2021}\natexlab{a}.
\newblock \showarticletitle{Rcd: Relation map driven cognitive diagnosis for
  intelligent education systems}. In \bibinfo{booktitle}{\emph{Proceedings of
  the 44th International ACM SIGIR Conference on Research and Development in
  Information Retrieval}}. \bibinfo{pages}{501--510}.
\newblock


\bibitem[Gao et~al\mbox{.}(2021b)]%
        {gao2021graph}
\bibfield{author}{\bibinfo{person}{Yang Gao}, \bibinfo{person}{Hong Yang},
  \bibinfo{person}{Peng Zhang}, \bibinfo{person}{Chuan Zhou}, {and}
  \bibinfo{person}{Yue Hu}.} \bibinfo{year}{2021}\natexlab{b}.
\newblock \showarticletitle{Graph neural architecture search}. In
  \bibinfo{booktitle}{\emph{Proceedings of the Twenty-Ninth International
  Conference on International Joint Conferences on Artificial Intelligence}}.
  \bibinfo{pages}{1403--1409}.
\newblock


\bibitem[Gurbanov et~al\mbox{.}(2016)]%
        {gurbanov2016modeling}
\bibfield{author}{\bibinfo{person}{Tural Gurbanov}, \bibinfo{person}{Francesco
  Ricci}, {and} \bibinfo{person}{Meinhard Ploner}.}
  \bibinfo{year}{2016}\natexlab{}.
\newblock \showarticletitle{Modeling and predicting user actions in recommender
  systems}. In \bibinfo{booktitle}{\emph{Proceedings of the 2016 Conference on
  User Modeling Adaptation and Personalization}}. \bibinfo{pages}{151--155}.
\newblock


\bibitem[He et~al\mbox{.}(2017)]%
        {he2017neural}
\bibfield{author}{\bibinfo{person}{Xiangnan He}, \bibinfo{person}{Lizi Liao},
  \bibinfo{person}{Hanwang Zhang}, \bibinfo{person}{Liqiang Nie},
  \bibinfo{person}{Xia Hu}, {and} \bibinfo{person}{Tat-Seng Chua}.}
  \bibinfo{year}{2017}\natexlab{}.
\newblock \showarticletitle{Neural collaborative filtering}. In
  \bibinfo{booktitle}{\emph{Proceedings of the 26th international conference on
  world wide web}}. \bibinfo{pages}{173--182}.
\newblock


\bibitem[Hochreiter(1998)]%
        {hochreiter1998vanishing}
\bibfield{author}{\bibinfo{person}{Sepp Hochreiter}.}
  \bibinfo{year}{1998}\natexlab{}.
\newblock \showarticletitle{The vanishing gradient problem during learning
  recurrent neural nets and problem solutions}.
\newblock \bibinfo{journal}{\emph{International Journal of Uncertainty,
  Fuzziness and Knowledge-Based Systems}} \bibinfo{volume}{6},
  \bibinfo{number}{02} (\bibinfo{year}{1998}), \bibinfo{pages}{107--116}.
\newblock


\bibitem[Hu et~al\mbox{.}(2023)]%
        {hu2023ptadisc}
\bibfield{author}{\bibinfo{person}{Liya Hu}, \bibinfo{person}{Zhiang Dong},
  \bibinfo{person}{Jingyuan Chen}, \bibinfo{person}{Guifeng Wang},
  \bibinfo{person}{Zhihua Wang}, \bibinfo{person}{Zhou Zhao}, {and}
  \bibinfo{person}{Fei Wu}.} \bibinfo{year}{2023}\natexlab{}.
\newblock \showarticletitle{PTADisc: A Cross-Course Dataset Supporting
  Personalized Learning in Cold-Start Scenarios}. In
  \bibinfo{booktitle}{\emph{Proceedings of the 37th Conference on Neural
  Information Processing Systems Datasets and Benchmarks Track}}.
\newblock


\bibitem[Huang et~al\mbox{.}(2020b)]%
        {huang2020loadaboost}
\bibfield{author}{\bibinfo{person}{Li Huang}, \bibinfo{person}{Yifeng Yin},
  \bibinfo{person}{Zeng Fu}, \bibinfo{person}{Shifa Zhang},
  \bibinfo{person}{Hao Deng}, {and} \bibinfo{person}{Dianbo Liu}.}
  \bibinfo{year}{2020}\natexlab{b}.
\newblock \showarticletitle{LoAdaBoost: Loss-based AdaBoost federated machine
  learning with reduced computational complexity on IID and non-IID intensive
  care data}.
\newblock \bibinfo{journal}{\emph{Plos one}} \bibinfo{volume}{15},
  \bibinfo{number}{4} (\bibinfo{year}{2020}), \bibinfo{pages}{e0230706}.
\newblock


\bibitem[Huang et~al\mbox{.}(2020a)]%
        {huang2020learning}
\bibfield{author}{\bibinfo{person}{Zhenya Huang}, \bibinfo{person}{Qi Liu},
  \bibinfo{person}{Yuying Chen}, \bibinfo{person}{Le Wu}, \bibinfo{person}{Keli
  Xiao}, \bibinfo{person}{Enhong Chen}, \bibinfo{person}{Haiping Ma}, {and}
  \bibinfo{person}{Guoping Hu}.} \bibinfo{year}{2020}\natexlab{a}.
\newblock \showarticletitle{Learning or forgetting? A dynamic approach for
  tracking the knowledge proficiency of students}.
\newblock \bibinfo{journal}{\emph{ACM Transactions on Information Systems
  (TOIS)}} \bibinfo{volume}{38}, \bibinfo{number}{2} (\bibinfo{year}{2020}),
  \bibinfo{pages}{1--33}.
\newblock


\bibitem[Hutter et~al\mbox{.}(2019)]%
        {hutter2019automated}
\bibfield{author}{\bibinfo{person}{Frank Hutter}, \bibinfo{person}{Lars
  Kotthoff}, {and} \bibinfo{person}{Joaquin Vanschoren}.}
  \bibinfo{year}{2019}\natexlab{}.
\newblock \bibinfo{booktitle}{\emph{Automated machine learning: methods,
  systems, challenges}}.
\newblock \bibinfo{publisher}{Springer Nature}.
\newblock


\bibitem[Ji et~al\mbox{.}(2019)]%
        {fedatt}
\bibfield{author}{\bibinfo{person}{Shaoxiong Ji}, \bibinfo{person}{Shirui Pan},
  \bibinfo{person}{Guodong Long}, \bibinfo{person}{Xue Li},
  \bibinfo{person}{Jing Jiang}, {and} \bibinfo{person}{Zi Huang}.}
  \bibinfo{year}{2019}\natexlab{}.
\newblock \showarticletitle{Learning private neural language modeling with
  attentive aggregation}. In \bibinfo{booktitle}{\emph{2019 International joint
  conference on neural networks (IJCNN)}}. IEEE, \bibinfo{pages}{1--8}.
\newblock


\bibitem[Jin et~al\mbox{.}(2023)]%
        {jin2023federated}
\bibfield{author}{\bibinfo{person}{Yaochu Jin}, \bibinfo{person}{Hangyu Zhu},
  \bibinfo{person}{Jinjin Xu}, {and} \bibinfo{person}{Yang Chen}.}
  \bibinfo{year}{2023}\natexlab{}.
\newblock \bibinfo{booktitle}{\emph{Federated Learning}}.
\newblock \bibinfo{publisher}{Springer}.
\newblock


\bibitem[Kairouz et~al\mbox{.}(2021)]%
        {kairouz2021advances}
\bibfield{author}{\bibinfo{person}{Peter Kairouz}, \bibinfo{person}{H~Brendan
  McMahan}, \bibinfo{person}{Brendan Avent}, \bibinfo{person}{Aur{\'e}lien
  Bellet}, \bibinfo{person}{Mehdi Bennis}, \bibinfo{person}{Arjun~Nitin
  Bhagoji}, \bibinfo{person}{Kallista Bonawitz}, \bibinfo{person}{Zachary
  Charles}, \bibinfo{person}{Graham Cormode}, \bibinfo{person}{Rachel
  Cummings}, {et~al\mbox{.}}} \bibinfo{year}{2021}\natexlab{}.
\newblock \showarticletitle{Advances and open problems in federated learning}.
\newblock \bibinfo{journal}{\emph{Foundations and trends{\textregistered} in
  machine learning}} \bibinfo{volume}{14}, \bibinfo{number}{1--2}
  (\bibinfo{year}{2021}), \bibinfo{pages}{1--210}.
\newblock


\bibitem[Koren et~al\mbox{.}(2009)]%
        {koren2009matrix}
\bibfield{author}{\bibinfo{person}{Yehuda Koren}, \bibinfo{person}{Robert
  Bell}, {and} \bibinfo{person}{Chris Volinsky}.}
  \bibinfo{year}{2009}\natexlab{}.
\newblock \showarticletitle{Matrix factorization techniques for recommender
  systems}.
\newblock \bibinfo{journal}{\emph{Computer}} \bibinfo{volume}{42},
  \bibinfo{number}{8} (\bibinfo{year}{2009}), \bibinfo{pages}{30--37}.
\newblock


\bibitem[Li et~al\mbox{.}(2022)]%
        {li2022hiercdf}
\bibfield{author}{\bibinfo{person}{Jiatong Li}, \bibinfo{person}{Fei Wang},
  \bibinfo{person}{Qi Liu}, \bibinfo{person}{Mengxiao Zhu},
  \bibinfo{person}{Wei Huang}, \bibinfo{person}{Zhenya Huang},
  \bibinfo{person}{Enhong Chen}, \bibinfo{person}{Yu Su}, {and}
  \bibinfo{person}{Shijin Wang}.} \bibinfo{year}{2022}\natexlab{}.
\newblock \showarticletitle{Hiercdf: A bayesian network-based hierarchical
  cognitive diagnosis framework}. In \bibinfo{booktitle}{\emph{Proceedings of
  the 28th ACM SIGKDD conference on knowledge discovery and data mining}}.
  \bibinfo{pages}{904--913}.
\newblock


\bibitem[Li et~al\mbox{.}(2020)]%
        {li2020review}
\bibfield{author}{\bibinfo{person}{Li Li}, \bibinfo{person}{Yuxi Fan},
  \bibinfo{person}{Mike Tse}, {and} \bibinfo{person}{Kuo-Yi Lin}.}
  \bibinfo{year}{2020}\natexlab{}.
\newblock \showarticletitle{A review of applications in federated learning}.
\newblock \bibinfo{journal}{\emph{Computers \& Industrial Engineering}}
  \bibinfo{volume}{149} (\bibinfo{year}{2020}), \bibinfo{pages}{106854}.
\newblock


\bibitem[Li et~al\mbox{.}(2023)]%
        {li2023graph}
\bibfield{author}{\bibinfo{person}{Qingyao Li}, \bibinfo{person}{Wei Xia},
  \bibinfo{person}{Li'ang Yin}, \bibinfo{person}{Jian Shen},
  \bibinfo{person}{Renting Rui}, \bibinfo{person}{Weinan Zhang},
  \bibinfo{person}{Xianyu Chen}, \bibinfo{person}{Ruiming Tang}, {and}
  \bibinfo{person}{Yong Yu}.} \bibinfo{year}{2023}\natexlab{}.
\newblock \showarticletitle{Graph enhanced hierarchical reinforcement learning
  for goal-oriented learning path recommendation}. In
  \bibinfo{booktitle}{\emph{Proceedings of the 32nd ACM International
  Conference on Information and Knowledge Management}}.
  \bibinfo{pages}{1318--1327}.
\newblock


\bibitem[Li et~al\mbox{.}(2019)]%
        {li2019convergence}
\bibfield{author}{\bibinfo{person}{Xiang Li}, \bibinfo{person}{Kaixuan Huang},
  \bibinfo{person}{Wenhao Yang}, \bibinfo{person}{Shusen Wang}, {and}
  \bibinfo{person}{Zhihua Zhang}.} \bibinfo{year}{2019}\natexlab{}.
\newblock \showarticletitle{On the convergence of fedavg on non-iid data}.
\newblock \bibinfo{journal}{\emph{arXiv preprint arXiv:1907.02189}}
  (\bibinfo{year}{2019}).
\newblock


\bibitem[Lin et~al\mbox{.}(2020a)]%
        {lin2020fedrec}
\bibfield{author}{\bibinfo{person}{Guanyu Lin}, \bibinfo{person}{Feng Liang},
  \bibinfo{person}{Weike Pan}, {and} \bibinfo{person}{Zhong Ming}.}
  \bibinfo{year}{2020}\natexlab{a}.
\newblock \showarticletitle{Fedrec: Federated recommendation with explicit
  feedback}.
\newblock \bibinfo{journal}{\emph{IEEE Intelligent Systems}}
  \bibinfo{volume}{36}, \bibinfo{number}{5} (\bibinfo{year}{2020}),
  \bibinfo{pages}{21--30}.
\newblock


\bibitem[Lin et~al\mbox{.}(2020b)]%
        {lin2020meta}
\bibfield{author}{\bibinfo{person}{Yujie Lin}, \bibinfo{person}{Pengjie Ren},
  \bibinfo{person}{Zhumin Chen}, \bibinfo{person}{Zhaochun Ren},
  \bibinfo{person}{Dongxiao Yu}, \bibinfo{person}{Jun Ma},
  \bibinfo{person}{Maarten~de Rijke}, {and} \bibinfo{person}{Xiuzhen Cheng}.}
  \bibinfo{year}{2020}\natexlab{b}.
\newblock \showarticletitle{Meta matrix factorization for federated rating
  predictions}. In \bibinfo{booktitle}{\emph{Proceedings of the 43rd
  International ACM SIGIR Conference on Research and Development in Information
  Retrieval}}. \bibinfo{pages}{981--990}.
\newblock


\bibitem[Liu et~al\mbox{.}(2024a)]%
        {liu2024fdkt}
\bibfield{author}{\bibinfo{person}{Fei Liu}, \bibinfo{person}{Chenyang Bu},
  \bibinfo{person}{Haotian Zhang}, \bibinfo{person}{Le Wu},
  \bibinfo{person}{Kui Yu}, {and} \bibinfo{person}{Xuegang Hu}.}
  \bibinfo{year}{2024}\natexlab{a}.
\newblock \showarticletitle{FDKT: Towards an interpretable deep knowledge
  tracing via fuzzy reasoning}.
\newblock \bibinfo{journal}{\emph{ACM Transactions on Information Systems}}
  \bibinfo{volume}{42}, \bibinfo{number}{5} (\bibinfo{year}{2024}),
  \bibinfo{pages}{1--26}.
\newblock


\bibitem[Liu et~al\mbox{.}(2023)]%
        {liu2023federated}
\bibfield{author}{\bibinfo{person}{Qi Liu}, \bibinfo{person}{Jinze Wu},
  \bibinfo{person}{Zhenya Huang}, \bibinfo{person}{Hao Wang},
  \bibinfo{person}{Yuting Ning}, \bibinfo{person}{Ming Chen},
  \bibinfo{person}{Enhong Chen}, \bibinfo{person}{Jinfeng Yi}, {and}
  \bibinfo{person}{Bowen Zhou}.} \bibinfo{year}{2023}\natexlab{}.
\newblock \showarticletitle{Federated user modeling from hierarchical
  information}.
\newblock \bibinfo{journal}{\emph{ACM Transactions on Information Systems}}
  \bibinfo{volume}{41}, \bibinfo{number}{2} (\bibinfo{year}{2023}),
  \bibinfo{pages}{1--33}.
\newblock


\bibitem[Liu et~al\mbox{.}(2022)]%
        {Liu2022Fairness}
\bibfield{author}{\bibinfo{person}{Shuchang Liu}, \bibinfo{person}{Yingqiang
  Ge}, \bibinfo{person}{Shuyuan Xu}, \bibinfo{person}{Yongfeng Zhang}, {and}
  \bibinfo{person}{Amelie Marian}.} \bibinfo{year}{2022}\natexlab{}.
\newblock \showarticletitle{Fairness-aware Federated Matrix Factorization}. In
  \bibinfo{booktitle}{\emph{Proceedings of the 16th ACM Conference on
  Recommender Systems}} (Seattle, WA, USA) \emph{(\bibinfo{series}{RecSys
  '22})}. \bibinfo{publisher}{Association for Computing Machinery},
  \bibinfo{address}{New York, NY, USA}, \bibinfo{pages}{168–178}.
\newblock
\showISBNx{9781450392785}
\urldef\tempurl%
\url{https://doi.org/10.1145/3523227.3546771}
\showURL{%
\tempurl}


\bibitem[Liu et~al\mbox{.}(2024b)]%
        {liu2024inductive}
\bibfield{author}{\bibinfo{person}{Shuo Liu}, \bibinfo{person}{Junhao Shen},
  \bibinfo{person}{Hong Qian}, {and} \bibinfo{person}{Aimin Zhou}.}
  \bibinfo{year}{2024}\natexlab{b}.
\newblock \showarticletitle{Inductive Cognitive Diagnosis for Fast Student
  Learning in Web-Based Intelligent Education Systems}. In
  \bibinfo{booktitle}{\emph{Proceedings of the ACM on Web Conference 2024}}.
  \bibinfo{pages}{4260--4271}.
\newblock


\bibitem[Ma et~al\mbox{.}(2022)]%
        {ma2022knowledge}
\bibfield{author}{\bibinfo{person}{Haiping Ma}, \bibinfo{person}{Manwei Li},
  \bibinfo{person}{Le Wu}, \bibinfo{person}{Haifeng Zhang},
  \bibinfo{person}{Yunbo Cao}, \bibinfo{person}{Xingyi Zhang}, {and}
  \bibinfo{person}{Xuemin Zhao}.} \bibinfo{year}{2022}\natexlab{}.
\newblock \showarticletitle{Knowledge-Sensed Cognitive Diagnosis for
  Intelligent Education Platforms}. In \bibinfo{booktitle}{\emph{Proceedings of
  the 31st ACM International Conference on Information \& Knowledge
  Management}}. \bibinfo{pages}{1451--1460}.
\newblock


\bibitem[McMahan et~al\mbox{.}(2017a)]%
        {fedavg}
\bibfield{author}{\bibinfo{person}{Brendan McMahan}, \bibinfo{person}{Eider
  Moore}, \bibinfo{person}{Daniel Ramage}, \bibinfo{person}{Seth Hampson},
  {and} \bibinfo{person}{Blaise~Aguera y Arcas}.}
  \bibinfo{year}{2017}\natexlab{a}.
\newblock \showarticletitle{Communication-efficient learning of deep networks
  from decentralized data}. In \bibinfo{booktitle}{\emph{Artificial
  intelligence and statistics}}. PMLR, \bibinfo{pages}{1273--1282}.
\newblock


\bibitem[McMahan et~al\mbox{.}(2017b)]%
        {mcmahan2017communication}
\bibfield{author}{\bibinfo{person}{Brendan McMahan}, \bibinfo{person}{Eider
  Moore}, \bibinfo{person}{Daniel Ramage}, \bibinfo{person}{Seth Hampson},
  {and} \bibinfo{person}{Blaise~Aguera y Arcas}.}
  \bibinfo{year}{2017}\natexlab{b}.
\newblock \showarticletitle{Communication-efficient learning of deep networks
  from decentralized data}. In \bibinfo{booktitle}{\emph{Artificial
  intelligence and statistics}}. PMLR, \bibinfo{pages}{1273--1282}.
\newblock


\bibitem[Nguyen et~al\mbox{.}(2021)]%
        {nguyen2021federated}
\bibfield{author}{\bibinfo{person}{Dinh~C Nguyen}, \bibinfo{person}{Ming Ding},
  \bibinfo{person}{Pubudu~N Pathirana}, \bibinfo{person}{Aruna Seneviratne},
  \bibinfo{person}{Jun Li}, {and} \bibinfo{person}{H~Vincent Poor}.}
  \bibinfo{year}{2021}\natexlab{}.
\newblock \showarticletitle{Federated learning for internet of things: A
  comprehensive survey}.
\newblock \bibinfo{journal}{\emph{IEEE Communications Surveys \& Tutorials}}
  \bibinfo{volume}{23}, \bibinfo{number}{3} (\bibinfo{year}{2021}),
  \bibinfo{pages}{1622--1658}.
\newblock


\bibitem[Perifanis and Efraimidis(2022)]%
        {perifanis2022federated}
\bibfield{author}{\bibinfo{person}{Vasileios Perifanis} {and}
  \bibinfo{person}{Pavlos~S Efraimidis}.} \bibinfo{year}{2022}\natexlab{}.
\newblock \showarticletitle{Federated neural collaborative filtering}.
\newblock \bibinfo{journal}{\emph{Knowledge-Based Systems}}
  \bibinfo{volume}{242} (\bibinfo{year}{2022}), \bibinfo{pages}{108441}.
\newblock


\bibitem[Pu et~al\mbox{.}(2024)]%
        {pu2024elakt}
\bibfield{author}{\bibinfo{person}{Yanjun Pu}, \bibinfo{person}{Fang Liu},
  \bibinfo{person}{Rongye Shi}, \bibinfo{person}{Haitao Yuan},
  \bibinfo{person}{Ruibo Chen}, \bibinfo{person}{Tianhao Peng}, {and}
  \bibinfo{person}{Wenjun Wu}.} \bibinfo{year}{2024}\natexlab{}.
\newblock \showarticletitle{ELAKT: Enhancing locality for attentive knowledge
  tracing}.
\newblock \bibinfo{journal}{\emph{ACM Transactions on Information Systems}}
  \bibinfo{volume}{42}, \bibinfo{number}{4} (\bibinfo{year}{2024}),
  \bibinfo{pages}{1--27}.
\newblock


\bibitem[Reckase(2009)]%
        {reckase2009multidimensional}
\bibfield{author}{\bibinfo{person}{Mark~D Reckase}.}
  \bibinfo{year}{2009}\natexlab{}.
\newblock \showarticletitle{Multidimensional item response theory models}.
\newblock In \bibinfo{booktitle}{\emph{Multidimensional item response theory}}.
  \bibinfo{publisher}{Springer}, \bibinfo{pages}{79--112}.
\newblock


\bibitem[Shafiq et~al\mbox{.}(2017)]%
        {MOOC}
\bibfield{author}{\bibinfo{person}{Huma Shafiq}, \bibinfo{person}{Zahid~Ashraf
  Wani}, \bibinfo{person}{Iram~Mukhtar Mahajan}, {and} \bibinfo{person}{Uzma
  Qadri}.} \bibinfo{year}{2017}\natexlab{}.
\newblock \showarticletitle{Courses beyond borders: A case study of MOOC
  platform Coursera}.
\newblock \bibinfo{journal}{\emph{Library Philosophy and Practice}}
  (\bibinfo{year}{2017}), \bibinfo{pages}{1--15}.
\newblock


\bibitem[Sun et~al\mbox{.}(2019)]%
        {sun2019evolving}
\bibfield{author}{\bibinfo{person}{Yanan Sun}, \bibinfo{person}{Bing Xue},
  \bibinfo{person}{Mengjie Zhang}, {and} \bibinfo{person}{Gary~G Yen}.}
  \bibinfo{year}{2019}\natexlab{}.
\newblock \showarticletitle{Evolving deep convolutional neural networks for
  image classification}.
\newblock \bibinfo{journal}{\emph{IEEE Transactions on Evolutionary
  Computation}} \bibinfo{volume}{24}, \bibinfo{number}{2}
  (\bibinfo{year}{2019}), \bibinfo{pages}{394--407}.
\newblock


\bibitem[Torre and J.(2009)]%
        {Torre2009DINA}
\bibfield{author}{\bibinfo{person}{D.~L. Torre} {and} \bibinfo{person}{J.}}
  \bibinfo{year}{2009}\natexlab{}.
\newblock \showarticletitle{DINA Model and Parameter Estimation: A Didactic}.
\newblock \bibinfo{journal}{\emph{Journal of Educational and Behavioral
  Statistics}} \bibinfo{volume}{34}, \bibinfo{number}{1}
  (\bibinfo{year}{2009}), \bibinfo{pages}{115--130}.
\newblock


\bibitem[Vuong et~al\mbox{.}(2025)]%
        {Tung2025Incorporating}
\bibfield{author}{\bibinfo{person}{Tung Vuong}, \bibinfo{person}{Pritom~Kumar
  Das}, {and} \bibinfo{person}{Tuukka Ruotsalo}.}
  \bibinfo{year}{2025}\natexlab{}.
\newblock \showarticletitle{Incorporating Cognitive Abilities into Web Search
  Re-ranking}.
\newblock \bibinfo{journal}{\emph{ACM Transactions on Information Systems}}
  (\bibinfo{date}{May} \bibinfo{year}{2025}).
\newblock
\showISSN{1046-8188}
\href{https://doi.org/10.1145/3736401}{doi:\nolinkurl{10.1145/3736401}}
\newblock
\shownote{Just Accepted}.


\bibitem[Wang et~al\mbox{.}(2024)]%
        {wang2024survey}
\bibfield{author}{\bibinfo{person}{Fei Wang}, \bibinfo{person}{Weibo Gao},
  \bibinfo{person}{Qi Liu}, \bibinfo{person}{Jiatong Li},
  \bibinfo{person}{Guanhao Zhao}, \bibinfo{person}{Zheng Zhang},
  \bibinfo{person}{Zhenya Huang}, \bibinfo{person}{Mengxiao Zhu},
  \bibinfo{person}{Shijin Wang}, \bibinfo{person}{Wei Tong}, {et~al\mbox{.}}}
  \bibinfo{year}{2024}\natexlab{}.
\newblock \showarticletitle{A Survey of Models for Cognitive Diagnosis: New
  Developments and Future Directions}.
\newblock \bibinfo{journal}{\emph{arXiv preprint arXiv:2407.05458}}
  (\bibinfo{year}{2024}).
\newblock


\bibitem[Wang et~al\mbox{.}(2020)]%
        {wang2020neural}
\bibfield{author}{\bibinfo{person}{Fei Wang}, \bibinfo{person}{Qi Liu},
  \bibinfo{person}{Enhong Chen}, \bibinfo{person}{Zhenya Huang},
  \bibinfo{person}{Yuying Chen}, \bibinfo{person}{Yu Yin}, \bibinfo{person}{Zai
  Huang}, {and} \bibinfo{person}{Shijin Wang}.}
  \bibinfo{year}{2020}\natexlab{}.
\newblock \showarticletitle{Neural cognitive diagnosis for intelligent
  education systems}. In \bibinfo{booktitle}{\emph{Proceedings of the 2020 AAAI
  Conference on Artificial Intelligence}}, Vol.~\bibinfo{volume}{34}.
  \bibinfo{pages}{6153--6161}.
\newblock


\bibitem[Wang et~al\mbox{.}(2022)]%
        {wang2022kancd}
\bibfield{author}{\bibinfo{person}{Fei Wang}, \bibinfo{person}{Qi Liu},
  \bibinfo{person}{Enhong Chen}, \bibinfo{person}{Zhenya Huang},
  \bibinfo{person}{Yu Yin}, \bibinfo{person}{Shijin Wang}, {and}
  \bibinfo{person}{Yu Su}.} \bibinfo{year}{2022}\natexlab{}.
\newblock \showarticletitle{NeuralCD: a general framework for cognitive
  diagnosis}.
\newblock \bibinfo{journal}{\emph{IEEE Transactions on Knowledge and Data
  Engineering}} (\bibinfo{year}{2022}).
\newblock


\bibitem[Wang et~al\mbox{.}(2023)]%
        {scd}
\bibfield{author}{\bibinfo{person}{Shanshan Wang}, \bibinfo{person}{Zhen Zeng},
  \bibinfo{person}{Xun Yang}, {and} \bibinfo{person}{Xingyi Zhang}.}
  \bibinfo{year}{2023}\natexlab{}.
\newblock \showarticletitle{Self-supervised Graph Learning for Long-tailed
  Cognitive Diagnosis}. In \bibinfo{booktitle}{\emph{Proceedings of the AAAI
  Conference on Artificial Intelligence}}, Vol.~\bibinfo{volume}{37}.
  \bibinfo{pages}{110--118}.
\newblock


\bibitem[Wang et~al\mbox{.}(2021)]%
        {wang2021using}
\bibfield{author}{\bibinfo{person}{Xinping Wang}, \bibinfo{person}{Caidie
  Huang}, \bibinfo{person}{Jinfang Cai}, {and} \bibinfo{person}{Liangyu Chen}.}
  \bibinfo{year}{2021}\natexlab{}.
\newblock \showarticletitle{Using knowledge concept aggregation towards
  accurate cognitive diagnosis}. In \bibinfo{booktitle}{\emph{Proceedings of
  the 30th ACM International Conference on Information \& Knowledge
  Management}}. \bibinfo{pages}{2010--2019}.
\newblock


\bibitem[Wu et~al\mbox{.}(2022)]%
        {wu2022federated}
\bibfield{author}{\bibinfo{person}{Chuhan Wu}, \bibinfo{person}{Fangzhao Wu},
  \bibinfo{person}{Lingjuan Lyu}, \bibinfo{person}{Tao Qi},
  \bibinfo{person}{Yongfeng Huang}, {and} \bibinfo{person}{Xing Xie}.}
  \bibinfo{year}{2022}\natexlab{}.
\newblock \showarticletitle{A federated graph neural network framework for
  privacy-preserving personalization}.
\newblock \bibinfo{journal}{\emph{Nature Communications}} \bibinfo{volume}{13},
  \bibinfo{number}{1} (\bibinfo{year}{2022}), \bibinfo{pages}{3091}.
\newblock


\bibitem[Wu et~al\mbox{.}(2021a)]%
        {wu2021federated}
\bibfield{author}{\bibinfo{person}{Jinze Wu}, \bibinfo{person}{Zhenya Huang},
  \bibinfo{person}{Qi Liu}, \bibinfo{person}{Defu Lian}, \bibinfo{person}{Hao
  Wang}, \bibinfo{person}{Enhong Chen}, \bibinfo{person}{Haiping Ma}, {and}
  \bibinfo{person}{Shijin Wang}.} \bibinfo{year}{2021}\natexlab{a}.
\newblock \showarticletitle{Federated deep knowledge tracing}. In
  \bibinfo{booktitle}{\emph{Proceedings of the 14th ACM international
  conference on web search and data mining}}. \bibinfo{pages}{662--670}.
\newblock


\bibitem[Wu et~al\mbox{.}(2021b)]%
        {hpfl}
\bibfield{author}{\bibinfo{person}{Jinze Wu}, \bibinfo{person}{Qi Liu},
  \bibinfo{person}{Zhenya Huang}, \bibinfo{person}{Yuting Ning},
  \bibinfo{person}{Hao Wang}, \bibinfo{person}{Enhong Chen},
  \bibinfo{person}{Jinfeng Yi}, {and} \bibinfo{person}{Bowen Zhou}.}
  \bibinfo{year}{2021}\natexlab{b}.
\newblock \showarticletitle{Hierarchical personalized federated learning for
  user modeling}. In \bibinfo{booktitle}{\emph{Proceedings of the Web
  Conference 2021}}. \bibinfo{pages}{957--968}.
\newblock


\bibitem[Wu et~al\mbox{.}(2020)]%
        {courserecommendation}
\bibfield{author}{\bibinfo{person}{Zhengyang Wu}, \bibinfo{person}{Ming Li},
  \bibinfo{person}{Yong Tang}, {and} \bibinfo{person}{Qingyu Liang}.}
  \bibinfo{year}{2020}\natexlab{}.
\newblock \showarticletitle{Exercise recommendation based on knowledge concept
  prediction}.
\newblock \bibinfo{journal}{\emph{Knowledge-Based Systems}}
  \bibinfo{volume}{210} (\bibinfo{year}{2020}), \bibinfo{pages}{106481}.
\newblock


\bibitem[Yang et~al\mbox{.}(2020)]%
        {yang2020federated}
\bibfield{author}{\bibinfo{person}{Liu Yang}, \bibinfo{person}{Ben Tan},
  \bibinfo{person}{Vincent~W Zheng}, \bibinfo{person}{Kai Chen}, {and}
  \bibinfo{person}{Qiang Yang}.} \bibinfo{year}{2020}\natexlab{}.
\newblock \showarticletitle{Federated recommendation systems}.
\newblock \bibinfo{journal}{\emph{Federated Learning: Privacy and Incentive}}
  (\bibinfo{year}{2020}), \bibinfo{pages}{225--239}.
\newblock


\bibitem[Yang et~al\mbox{.}(2024)]%
        {yang2024evolutionaryCD}
\bibfield{author}{\bibinfo{person}{Shangshang Yang}, \bibinfo{person}{Haiping
  Ma}, \bibinfo{person}{Ying Bi}, \bibinfo{person}{Ye Tian},
  \bibinfo{person}{Limiao Zhang}, \bibinfo{person}{Yaochu Jin}, {and}
  \bibinfo{person}{Xingyi Zhang}.} \bibinfo{year}{2024}\natexlab{}.
\newblock \showarticletitle{An Evolutionary Multi-Objective Neural Architecture
  Search Approach to Advancing Cognitive Diagnosis in Intelligent Education}.
\newblock \bibinfo{journal}{\emph{IEEE Transactions on Evolutionary
  Computation}} (\bibinfo{year}{2024}).
\newblock


\bibitem[Yang et~al\mbox{.}(2023a)]%
        {yang2023designing}
\bibfield{author}{\bibinfo{person}{Shangshang Yang}, \bibinfo{person}{Haiping
  Ma}, \bibinfo{person}{Cheng Zhen}, \bibinfo{person}{Ye Tian},
  \bibinfo{person}{Limiao Zhang}, \bibinfo{person}{Yaochu Jin}, {and}
  \bibinfo{person}{Xingyi Zhang}.} \bibinfo{year}{2023}\natexlab{a}.
\newblock \showarticletitle{Designing novel cognitive diagnosis models via
  evolutionary multi-objective neural architecture search}.
\newblock \bibinfo{journal}{\emph{arXiv preprint arXiv:2307.04429}}
  (\bibinfo{year}{2023}).
\newblock


\bibitem[Yang et~al\mbox{.}(2022)]%
        {yang2022accelerating}
\bibfield{author}{\bibinfo{person}{Shangshang Yang}, \bibinfo{person}{Ye Tian},
  \bibinfo{person}{Xiaoshu Xiang}, \bibinfo{person}{Shichen Peng}, {and}
  \bibinfo{person}{Xingyi Zhang}.} \bibinfo{year}{2022}\natexlab{}.
\newblock \showarticletitle{Accelerating Evolutionary Neural Architecture
  Search via Multifidelity Evaluation}.
\newblock \bibinfo{journal}{\emph{IEEE Transactions on Cognitive and
  Developmental Systems}} \bibinfo{volume}{14}, \bibinfo{number}{4}
  (\bibinfo{year}{2022}), \bibinfo{pages}{1778--1792}.
\newblock


\bibitem[Yang et~al\mbox{.}(2023b)]%
        {yang2023cognitive}
\bibfield{author}{\bibinfo{person}{Shangshang Yang}, \bibinfo{person}{Haoyu
  Wei}, \bibinfo{person}{Haiping Ma}, \bibinfo{person}{Ye Tian},
  \bibinfo{person}{Xingyi Zhang}, \bibinfo{person}{Yunbo Cao}, {and}
  \bibinfo{person}{Yaochu Jin}.} \bibinfo{year}{2023}\natexlab{b}.
\newblock \showarticletitle{Cognitive diagnosis-based personalized exercise
  group assembly via a multi-objective evolutionary algorithm}.
\newblock \bibinfo{journal}{\emph{IEEE Transactions on Emerging Topics in
  Computational Intelligence}} (\bibinfo{year}{2023}).
\newblock


\bibitem[Yang et~al\mbox{.}(2023c)]%
        {yang2023evolutionaryGCD}
\bibfield{author}{\bibinfo{person}{Shangshang Yang}, \bibinfo{person}{Cheng
  Zhen}, \bibinfo{person}{Ye Tian}, \bibinfo{person}{Haiping Ma},
  \bibinfo{person}{Yuanchao Liu}, \bibinfo{person}{Panpan Zhang}, {and}
  \bibinfo{person}{Xingyi Zhang}.} \bibinfo{year}{2023}\natexlab{c}.
\newblock \showarticletitle{Evolutionary Multi-Objective Neural Architecture
  Search for Generalized Cognitive Diagnosis Models}. In
  \bibinfo{booktitle}{\emph{Proceedings of the 5th International Conference on
  Data-driven Optimization of Complex Systems}}. IEEE, \bibinfo{pages}{1--10}.
\newblock


\bibitem[Yang et~al\mbox{.}(2023d)]%
        {yang2023evolutionary}
\bibfield{author}{\bibinfo{person}{Shangshang Yang}, \bibinfo{person}{Cheng
  Zhen}, \bibinfo{person}{Ye Tian}, \bibinfo{person}{Haiping Ma},
  \bibinfo{person}{Yuanchao Liu}, \bibinfo{person}{Panpan Zhang}, {and}
  \bibinfo{person}{Xingyi Zhang}.} \bibinfo{year}{2023}\natexlab{d}.
\newblock \showarticletitle{Evolutionary Multi-Objective Neural Architecture
  Search for Generalized Cognitive Diagnosis Models}. In
  \bibinfo{booktitle}{\emph{2023 5th International Conference on Data-driven
  Optimization of Complex Systems (DOCS)}}. IEEE, \bibinfo{pages}{1--10}.
\newblock


\bibitem[Yang et~al\mbox{.}(2018)]%
        {yang2018applied}
\bibfield{author}{\bibinfo{person}{Timothy Yang}, \bibinfo{person}{Galen
  Andrew}, \bibinfo{person}{Hubert Eichner}, \bibinfo{person}{Haicheng Sun},
  \bibinfo{person}{Wei Li}, \bibinfo{person}{Nicholas Kong},
  \bibinfo{person}{Daniel Ramage}, {and} \bibinfo{person}{Fran{\c{c}}oise
  Beaufays}.} \bibinfo{year}{2018}\natexlab{}.
\newblock \showarticletitle{Applied federated learning: Improving google
  keyboard query suggestions}.
\newblock \bibinfo{journal}{\emph{arXiv preprint arXiv:1812.02903}}
  (\bibinfo{year}{2018}).
\newblock


\bibitem[Yang et~al\mbox{.}(2019)]%
        {yang2019ffd}
\bibfield{author}{\bibinfo{person}{Wensi Yang}, \bibinfo{person}{Yuhang Zhang},
  \bibinfo{person}{Kejiang Ye}, \bibinfo{person}{Li Li}, {and}
  \bibinfo{person}{Cheng-Zhong Xu}.} \bibinfo{year}{2019}\natexlab{}.
\newblock \showarticletitle{Ffd: A federated learning based method for credit
  card fraud detection}. In \bibinfo{booktitle}{\emph{Big Data--BigData 2019:
  8th International Congress, Held as Part of the Services Conference
  Federation, SCF 2019, San Diego, CA, USA, June 25--30, 2019, Proceedings 8}}.
  Springer, \bibinfo{pages}{18--32}.
\newblock


\bibitem[Yu et~al\mbox{.}(2020)]%
        {yu2020fairness}
\bibfield{author}{\bibinfo{person}{Han Yu}, \bibinfo{person}{Zelei Liu},
  \bibinfo{person}{Yang Liu}, \bibinfo{person}{Tianjian Chen},
  \bibinfo{person}{Mingshu Cong}, \bibinfo{person}{Xi Weng},
  \bibinfo{person}{Dusit Niyato}, {and} \bibinfo{person}{Qiang Yang}.}
  \bibinfo{year}{2020}\natexlab{}.
\newblock \showarticletitle{A fairness-aware incentive scheme for federated
  learning}. In \bibinfo{booktitle}{\emph{Proceedings of the AAAI/ACM
  Conference on AI, Ethics, and Society}}. \bibinfo{pages}{393--399}.
\newblock


\bibitem[Zhang et~al\mbox{.}(2023)]%
        {zhang2023dual}
\bibfield{author}{\bibinfo{person}{Chunxu Zhang}, \bibinfo{person}{Guodong
  Long}, \bibinfo{person}{Tianyi Zhou}, \bibinfo{person}{Peng Yan},
  \bibinfo{person}{Zijian Zhang}, \bibinfo{person}{Chengqi Zhang}, {and}
  \bibinfo{person}{Bo Yang}.} \bibinfo{year}{2023}\natexlab{}.
\newblock \showarticletitle{Dual personalization on federated recommendation}.
\newblock \bibinfo{journal}{\emph{arXiv preprint arXiv:2301.08143}}
  (\bibinfo{year}{2023}).
\newblock


\bibitem[Zhang et~al\mbox{.}(2024a)]%
        {zhang2024gpfedrec}
\bibfield{author}{\bibinfo{person}{Chunxu Zhang}, \bibinfo{person}{Guodong
  Long}, \bibinfo{person}{Tianyi Zhou}, \bibinfo{person}{Zijian Zhang},
  \bibinfo{person}{Peng Yan}, {and} \bibinfo{person}{Bo Yang}.}
  \bibinfo{year}{2024}\natexlab{a}.
\newblock \showarticletitle{GPFedRec: Graph-Guided Personalization for
  Federated Recommendation}. In \bibinfo{booktitle}{\emph{Proceedings of the
  30th ACM SIGKDD Conference on Knowledge Discovery and Data Mining}}.
  \bibinfo{pages}{4131--4142}.
\newblock


\bibitem[Zhang et~al\mbox{.}(2021)]%
        {zhang2021survey}
\bibfield{author}{\bibinfo{person}{Chen Zhang}, \bibinfo{person}{Yu Xie},
  \bibinfo{person}{Hang Bai}, \bibinfo{person}{Bin Yu},
  \bibinfo{person}{Weihong Li}, {and} \bibinfo{person}{Yuan Gao}.}
  \bibinfo{year}{2021}\natexlab{}.
\newblock \showarticletitle{A survey on federated learning}.
\newblock \bibinfo{journal}{\emph{Knowledge-Based Systems}}
  \bibinfo{volume}{216} (\bibinfo{year}{2021}), \bibinfo{pages}{106775}.
\newblock


\bibitem[Zhang et~al\mbox{.}(2024b)]%
        {zhang2024towards}
\bibfield{author}{\bibinfo{person}{Zheng Zhang}, \bibinfo{person}{Wei Song},
  \bibinfo{person}{Qi Liu}, \bibinfo{person}{Qingyang Mao},
  \bibinfo{person}{Yiyan Wang}, \bibinfo{person}{Weibo Gao},
  \bibinfo{person}{Zhenya Huang}, \bibinfo{person}{Shijin Wang}, {and}
  \bibinfo{person}{Enhong Chen}.} \bibinfo{year}{2024}\natexlab{b}.
\newblock \showarticletitle{Towards accurate and fair cognitive diagnosis via
  monotonic data augmentation}.
\newblock \bibinfo{journal}{\emph{Advances in Neural Information Processing
  Systems}}  \bibinfo{volume}{37} (\bibinfo{year}{2024}),
  \bibinfo{pages}{47767--47789}.
\newblock


\bibitem[Zhu et~al\mbox{.}(2021)]%
        {zhu2021federated}
\bibfield{author}{\bibinfo{person}{Hangyu Zhu}, \bibinfo{person}{Jinjin Xu},
  \bibinfo{person}{Shiqing Liu}, {and} \bibinfo{person}{Yaochu Jin}.}
  \bibinfo{year}{2021}\natexlab{}.
\newblock \showarticletitle{Federated learning on non-IID data: A survey}.
\newblock \bibinfo{journal}{\emph{Neurocomputing}}  \bibinfo{volume}{465}
  (\bibinfo{year}{2021}), \bibinfo{pages}{371--390}.
\newblock


\bibitem[Zoph and Le(2016)]%
        {zoph2016neural}
\bibfield{author}{\bibinfo{person}{Barret Zoph} {and} \bibinfo{person}{Quoc~V
  Le}.} \bibinfo{year}{2016}\natexlab{}.
\newblock \showarticletitle{Neural architecture search with reinforcement
  learning}.
\newblock \bibinfo{journal}{\emph{arXiv preprint arXiv:1611.01578}}
  (\bibinfo{year}{2016}).
\newblock


\end{thebibliography}

\end{document}